\documentclass[letterpaper, 10pt, conference]{ieeeconf}      % Use this line for a4 paper

\usepackage{graphics} % for pdf, bitmapped graphics files
\usepackage{epsfig} % for postscript graphics files
\usepackage{mathptmx} % assumes new font selection scheme installed
\usepackage{times} % assumes new font selection scheme installed
\usepackage{amsmath} % assumes amsmath package installed
\usepackage{amssymb}  % assumes amsmath package installed
\usepackage{mathtools}
\usepackage{graphicx}
\usepackage[font=small,labelsep=period]{caption}
\usepackage{calrsfs}
\DeclareMathAlphabet{\pazocal}{OMS}{zplm}{m}{n}

\usepackage{epsfig} % for postscript graphics files

\usepackage{subfloat}
\usepackage{epstopdf}
\usepackage{pict2e}
\usepackage{tikz}
\usetikzlibrary{shapes,arrows,decorations.markings}
\usepackage{bm}
\usepackage{adjustbox}
\usetikzlibrary{positioning}
\usepackage{lipsum}% http://ctan.org/pkg/lipsum %% WHY :
\usepackage{arydshln}
\usepackage[shortcuts,acronym]{glossaries}
\makeglossaries % generates the acronym list
\usepackage{multicol}
\usepackage{relsize}
\usepackage{tikz-3dplot}
\usepackage[english]{babel}
\usepackage{filecontents}
\usepackage{gensymb}
\usepackage{float}
\usepackage{caption,subcaption}

\usepackage[printwatermark]{xwatermark}
\usepackage{xcolor}
\usepackage{graphicx}
\usepackage{lipsum}

\IEEEoverridecommandlockouts                              
\title{\LARGE \bf
Safe Robotic Grasping: Minimum Impact-Force Grasp Selection 
}

\author{IEEE copyright notice  (To be appeared in IROS 2017) \\
Nikos Mavrakis, Amir M. Ghalamzan E. and Rustam Stolkin % <-this % stops a space
\thanks{All authors are with University of Birmingham, UK. \{nxm504, A.Ghalamzanesfahani, R.Stolkin\} @bham.ac.uk. This project was funded by EU H2020 RoMaNS, 645582, and EPSRC EP/M026477/1. Stolkin was supported by a Royal Society Industry Fellowship.}% <-this % stops a space
}

\begin{document}

\maketitle
\thispagestyle{empty}
\pagestyle{empty}

%%%%%%%%%%%%%%%%%%%%%%%%%%%%%%%%%%%%%%%%%%%%%%%%%%%%%%%%%%%%%%%%%%%%%%%%%%%%%%%%
\begin{abstract}
This paper addresses the problem of selecting from a choice of possible grasps, so that impact forces will be minimised if a collision occurs while the robot is moving the grasped object along a post-grasp trajectory. Such considerations are important for safety in human-robot interaction, where even a certified ``human-safe'' (e.g. compliant) arm may become hazardous once it grasps and begins moving an object, which may have significant mass, sharp edges or other dangers. Additionally, minimising collision forces is critical to preserving longevity of robots which operate in uncertain and hazardous environments, e.g. robots deployed for nuclear decommissioning, where removing a damaged robot from a contaminated zone for repairs may be extremely difficult and costly. Also, unwanted collisions between a robot and critical infrastructure (e.g. pipework) in such high-consequence environments can be disastrous. In this paper we investigate how the safety of the post-grasp motion can be considered during the pre-grasp approach phase, so that the selected grasp is optimal in terms applying minimum impact forces if a collision occurs during a desired post-grasp manipulation. We build on the methods of augmented robot-object dynamics models and ``effective mass''
and propose a method for combining these concepts with modern grasp and trajectory planners, to enable the robot to achieve a grasp which maximises the safety of the post-grasp trajectory, by minimising potential collision forces. We demonstrate the effectiveness of our approach through several experiments with both simulated and real robots.

\end{abstract}

%%%%%%%%%%%%%%%%%%%%%%%%%%%%%%%%%%%%%%%%%%%%%%%%%%%%%%%%%%%%%%%%%%%%%%%%%%%%%%%%
%%%%%%%%%%%%%%%%%%%%%%%%%%%%%%%%%%%%%%%%%%%%%%%%%%%%%%%%%%%%%%%%%%%%%%%%%%%%%%%%%%
%
%
%
% Amir checked first time, 010320170618 
%
%
%
%%%%%%%%%%%%%%%%%%%%%%%%%%%%%%%%%%%%%%%%%%%%%%%%%%%%%%%%%%%%%%%%%%%%%%%%%%%%%%%%%%%

\section{Introduction}
The fundamental reason for grasping an object is so that you can move it to another location. However, the vast majority of robotic grasping literature has focused on choosing appropriate finger positions for grasping, without considering the desired post-grasp manipulative trajectory, which will be needed to move the grasped object in a useful way.

\begin{figure}[tb!]
\centering
\begin{subfigure}[b]{0.43\linewidth}
 \includegraphics[width = 1\linewidth]{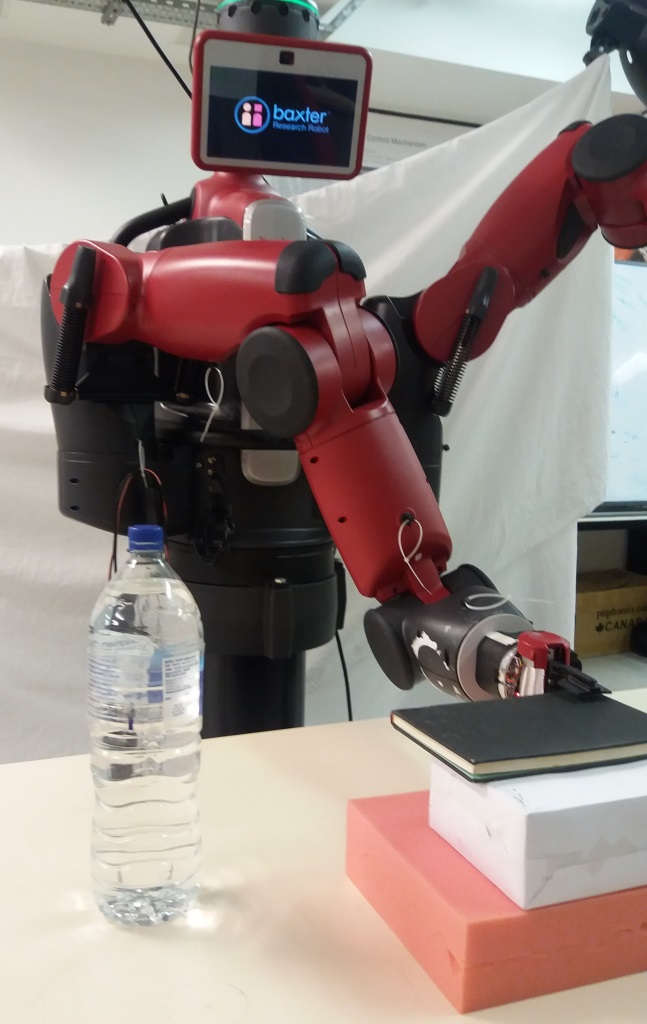}
   \caption{ \label{Fig:RealConfig}}
\end{subfigure}
  \quad
\begin{subfigure}[b]{.4\linewidth}
 \includegraphics[width = 1\linewidth]{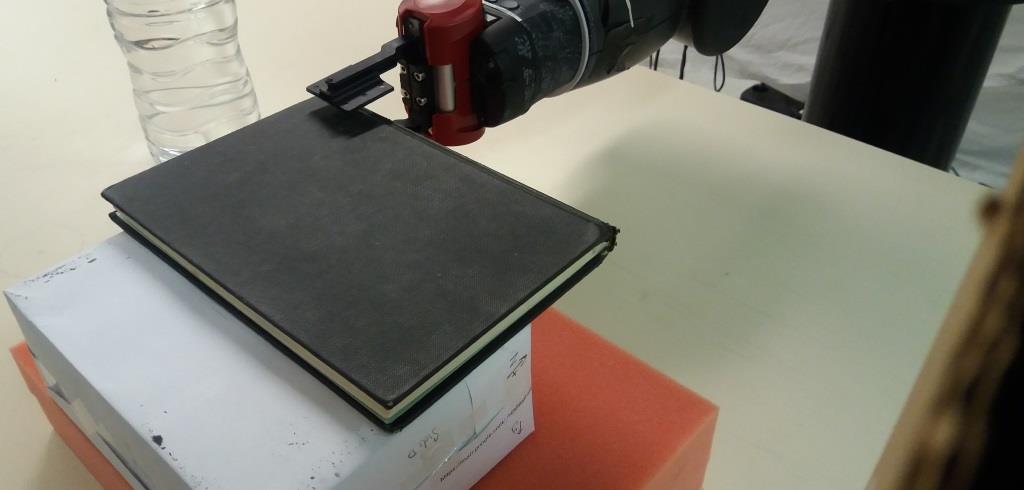}
  \caption{  \label{Fig:G1}}
    \vspace{0.5pt}
\includegraphics[width = 1\linewidth]{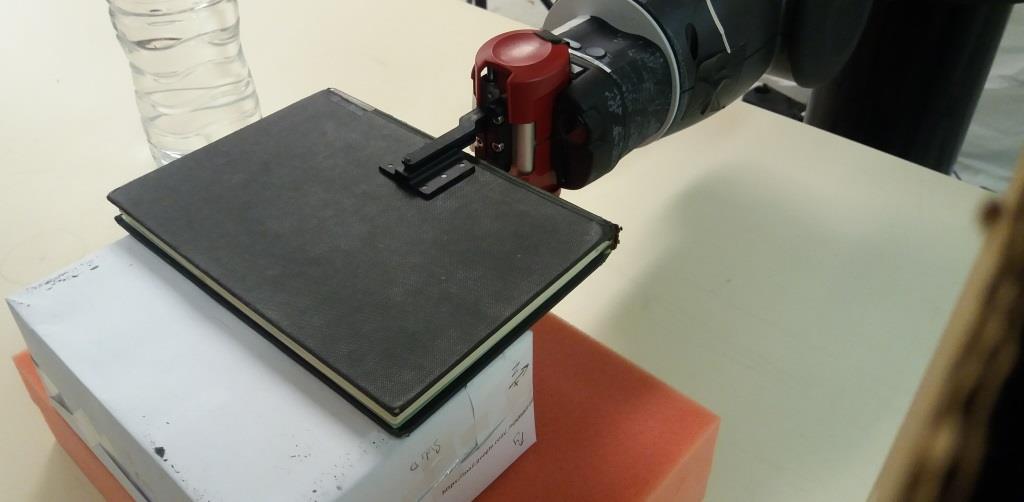}
   \caption{ \label{Fig:G2}  }
    \vspace{0.5pt}
\includegraphics[width = 1\linewidth]{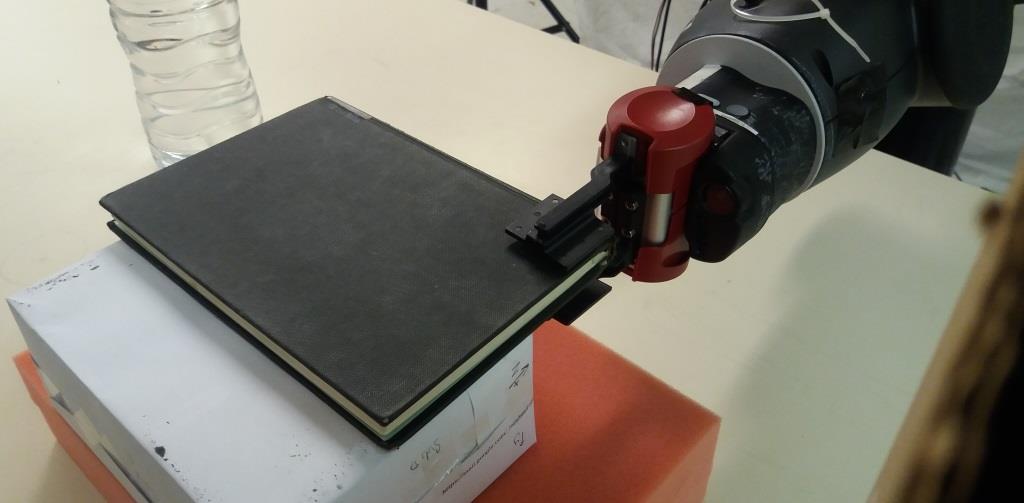}
  \caption{  \label{Fig:G3}}
\end{subfigure}
\caption{Pick and place experiment with a real robot. The robot is tasked with grasping and then moving the book (which has same dimensions, mass and task as used in simulation experiments described later in this paper). During the post-grasp manipulation, a collision occurs between the grasped object and an obstacle (the water bottle) and force values in the end-effector are measured. This paper is concerned with enabling the robot to choose from several feasible grasps, to minimise impact forces during such post-grasp collisions. In this experiment, the impact forces are measured for each possible grasp on the book. (\subref{Fig:RealConfig}) Overall experimental configuration. (\subref{Fig:G1}) First grasp choice. (\subref{Fig:G2}) Second grasp choice. (\subref{Fig:G3}) Third grasp choice.
% The purpose of this experiment is to check whether the real force measurements are in accordance with the simulated ones.
}\label{fig:RealExp}
\end{figure}

Because robotic grasping and robot arm trajectory planning are both complex problems in their own right, these problems have mostly been studied separately from each other. For example, consider the common robotic task of pick-and-place, shown in Fig.~\ref{fig:RealExp}, in which the robot is tasked with grasping the book and then moving it to a desired destination. Methods for reaching the object and grasping it may include predicting stable grasps from visual information \cite{kopicki2015one}, while methods for planning the post-grasp trajectory might include learning trajectories from demonstration~\cite{osa2017guiding} or optimising a trajectory using numerical methods~\cite{ratliff2009chomp}.

Unfortunately, many grasps suggested by a grasp planning algorithm might be sub-optimal (or even impossible) for achieving the desired post-grasp motion. Our previous work showed how poor choices of grasp may make the desired movement of the grasped object kinematically infeasible for the arm to achieve \cite{Ghalamzan2016}, or can result in unnecessarily high joint torques during post-grasp manipulations \cite{Mavrakis2016}.

This paper extends our previous ideas on ``task-informed grasp selection'' \cite{Ghalamzan2016,Mavrakis2016}, by proposing a method for choosing grasps which maximise safety during post-grasp manipulations, in dynamic and uncertain environments where collisions are both likely and also safety-critical. Motivating applications include human-robot collaborative working, or remote manipulation in highly cluttered and high-consequence environments such as nuclear decommissioning and robotic surgery \cite{wilkening2017development, alambeigi2011simulation}.

% Apart from automated plants, post-grasp manipulations are likely to occur in environments where the robots collaborate with humans, such as out of the cage in small to medium size industries as well as educational and office environments. As a result, a robot needs to be acting in a way that its post-grasp manipulations are safe when a human is present and therefore, safety is an important property to be satisfied by the provided robot's architecture. In this paper, we study how we can evaluate grasp safety during grasp selection, to select the grasp corresponding with the highest safety measure over the post-grasp trajectory.

% Many robotic tasks involve object grasping and manipulation. 
% However, the tasks of approaching an arbitrary object and generating a grasp, and then performing a desired manipulative motion are very complex. Hence, they are mostly studied separately. For instance, consider the common robotic task of pick-and-place, shown in Fig.~\ref{fig:RealExp}, in which he robot has to lift the book and place it in a desired target point. 
%

% it may be the case that the synthesised grasps are not suitable for motion planning, e.g. they bring the manipulator close to the joint limits. In the framework of a robot handling an object to execute a predefined task, there is a number of post-grasp properties that can be optimized. Our previous works \cite{Mavrakis2016}\cite{Ghalamzan2016} showed the cases of robots selecting the grasps that optimize post-grasp manipulability and minimize joint torques. 

Making robotic manipulation safer for humans and surroundings is a domain that is receiving increasing attention. Many approaches have been proposed to tackle this problem. For example, Ikuta et. al~\cite{Ikuta2003} define three categories of robot safety measures, including collision avoidance, impact minimisation and post-impact suppression methods. 
A large proportion of the research literature has been devoted to the collision avoidance category, e.g. \cite{Flacco2012} and \cite{Beetz2015}. In contrast, in this paper we focus on the category of impact minimisation.

Early work on impact minimisation includes \cite{Walker1990}, where the authors described the impact force, delivered by the robot's end-effector, as a function of its velocity, type of impact, impact direction and manipulator posture. They use the manipulator's redundancy to alter the posture in order to minimise the impact force. Impact force optimisation is also studied in \cite{Lin1995}, where the authors designed a compliance control method to reduce post-collision bouncing effects. In \cite{Walker1994} and \cite{Barcio1994}, the authors defined the impact ellipsoid of a manipulator as the variation of the end-effector impact force w.r.t. variations in the joint velocities. Kim et. al~\cite{Kim2000} extended those notations and proposed new impact force measures related to the robot's directional velocity. Furthermore, Heinzmann et al~\cite{Heinzmann2003} defined  \textit{impact potential} as a quantification of the maximum impact force a robot can exert in a collision with a stationary object. They implemented controllers to directly control the impact potential. In \cite{hu2016pre} the impact ellipsoid is  expressed as a series of inertia quasi-ellipsoids for a space robot and object model, and a pre-impact configuration is  designed to inflict minimum impact forces before grasping an object. The ideas of \cite{hu2016pre} are related to the work presented in this paper, in the sense that \cite{hu2016pre} uses an object's inertial properties to minimise impact \textit{before} grasping. However, our paper is concerned with understanding inertial properties to minimise impact \textit{after} grasping. 

Khatib et. al~\cite{khatib1995} analysed the inertial properties of robotic manipulators and introduced the terms \textit{effective mass and inertia}. These terms were used to describe the mass and inertia felt by the environment during a collision with the robot's end-effector. In \cite{Kang2010}, the effective mass was minimised for the case of a mobile manipulator, by using both the mobile platform's and the manipulator's kinematic redundancies. Haddadin et. al~\cite{Haddadin2012} used variable effective mass, inertia, and robot velocity to approximate collision with human tissues. They generated a database that describes the effects of different collision configurations on human tissue, embedding injury knowledge in the robot's motion planning and control systems.
In \cite{Baena2014} and \cite{Petersen2016}, a surgery robot's redundancy was used to follow a minimum effective mass and inertia trajectory while under the surgeons control. Another strategy in impact minimisation is presented in \cite{Rossi2015}, where the authors designed a robot path controller to constrain the dissipated energy in case of inelastic collision. Finally,  Ragaglia et. al~\cite{Ragaglia2014} proposed an integrating approach. The proposed approach combines visual and sensory input, as well as minimisation of reflected mass and robot velocity regulation to estimate a severity index when a person is nearby.

While the above-mentioned approaches explore ways of making the robot safer, they do not consider the safety of the manipulator while it is holding and manipulating an object. Nonetheless, robots are intended to manipulate objects. Holding an object alters the dynamics of the manipulator and consequently affects the impact forces experienced in a collision. The impact forces on the end-effector will change when holding the same object with different grasp poses, because each grasping pose corresponds to a different transformation of the object's inertia tensor with respect to the grasp frame. 

Although there have been many studies on robotic safety during manipulation, they consider the problem of safety during manipulative motion only. In contrast, our work is different in that we consider safety \textit{prior to grasping}, and incorporate notions of post-grasp safety into the grasp selection process. To the best of our knowledge, the effects of different grasp pose choices on post-grasp impact force and robot safety have not previously been studied, and previous grasping methods have not incorporated metrics of post-grasp safety into the grasp planning process.
%We use the objects inertial properties for the grasp selection, and our proposed approach allows a robot to chose a grasp with the minimum impact force in case of a collision with the robot's end-effector.

The main contributions of this paper are as follows. The paper provides an analysis of the effect of different object grasping poses on the augmented robot-object dynamics, as well as evaluating their performance in the case of post-grasp collisions. We develop a simple but foundational methodology in an under-explored aspect of robotic safety. We propose and evaluate a novel grasp selection criterion for post-grasp manipulation, that enables a robot to choose a grasp which minimises the resulting impact force in the event of a post-grasp collision with the robot's end-effector. Note that this post-grasp safety criterion can readily be combined with ``graspability'' or ``grasp-likelihood'' metrics, proposed in recent grasping literature such as \cite{kopicki2015one}, in order to choose grasps which mutually optimise both the success of the grasp and also the safety of the post-grasp manipulation.

To show the effectiveness of our approach, and to evaluate our hypotheses, we present the results of a series of empirical experiments, which evaluate and illustrate how the impact force is related to both the object's inertial properties and the robot's choice of grasp .

%%%%%%%%%%%%%%%%%%%%%%%%%%%%%%%%%%%%%%%%%%%%%%%%%%%%%%%%%%%%%%%%%%%%%%%%%%%%%%%%%%%%%%%%%%%%%%%%%
%
%
%
% Amir Checked PROBLEM FORMULATION SECTION in the following : 010320170336
%
%
%
%%%%%%%%%%%%%%%%%%%%%%%%%%%%%%%%%%%%%%%%%%%%%%%%%%%%%%%%%%%%%%%%%%%%%%%%%%%%%%%%%%%%%%%%%%%%%%%%%

\section{Problem Formulation}
\begin{figure}[t!]
\centering
     \scalebox{0.8}{\tdplotsetmaincoords{70}{110}
\begin{tikzpicture}[tdplot_main_coords,scale = 0.7]

\draw [very thick,dashed,blue] plot [smooth, tension=1] coordinates { (0,0,0) (5,5,2) (8,9,7) (12,14,9) (15,18,8)};

\draw[thick,->] (0,0,0) -- (1,0,0) node[anchor=north east]{$x_0$};
\draw[thick,->] (0,0,0) -- (0,1,0) node[anchor=north west]{$y_0$};
\draw[thick,->] (0,0,0) -- (0,0,1) node[anchor=south]{$z_0$};

\tdplotsetrotatedcoords{10}{10}{80}
\coordinate (Shift) at (5,5,2);
\tdplotsetrotatedcoordsorigin{(Shift)}

\draw[thick,->,tdplot_rotated_coords] (0,0,0) -- (1,0,0) node[anchor=north east]{$x_1$};
\draw[thick,->,tdplot_rotated_coords] (0,0,0) -- (0,1,0) node[anchor=north west]{$y_1$};
\draw[thick,->,tdplot_rotated_coords] (0,0,0) -- (0,0,1) node[anchor=south]{$z_1$};

\tdplotsetrotatedcoords{30}{10}{50}
\coordinate (Shift) at (8,9,7);
\tdplotsetrotatedcoordsorigin{(Shift)}

\draw[thick,->,tdplot_rotated_coords] (0,0,0) -- (1,0,0) node[anchor=north east]{$x_2$};
\draw[thick,->,tdplot_rotated_coords] (0,0,0) -- (0,1,0) node[anchor=north west]{$y_2$};
\draw[thick,->,tdplot_rotated_coords] (0,0,0) -- (0,0,1) node[anchor=south]{$z_2$};

\tdplotsetrotatedcoords{10}{10}{-10}
\coordinate (Shift) at (12,14,9);
\tdplotsetrotatedcoordsorigin{(Shift)}

\draw[thick,->,tdplot_rotated_coords] (0,0,0) -- (1,0,0) node[anchor=north east]{$x_3$};
\draw[thick,->,tdplot_rotated_coords] (0,0,0) -- (0,1,0) node[anchor=north west]{$y_3$};
\draw[thick,->,tdplot_rotated_coords] (0,0,0) -- (0,0,1) node[anchor=south]{$z_3$};

\tdplotsetrotatedcoords{-30}{10}{10}
\coordinate (Shift) at (15,18,8);
\tdplotsetrotatedcoordsorigin{(Shift)}

\draw[thick,->,tdplot_rotated_coords] (0,0,0) -- (1,0,0) node[anchor=north east]{$x_f$};
\draw[thick,->,tdplot_rotated_coords] (0,0,0) -- (0,1,0) node[anchor=north west]{$y_f$};
\draw[thick,->,tdplot_rotated_coords] (0,0,0) -- (0,0,1) node[anchor=south]{$z_f$};
\end{tikzpicture}}  

\caption{End-effector trajectory example. The desired trajectory, which the robot is commanded to follow post-grasp, is considered known prior to grasping. The robot needs to move from an initial pose to a target pose, $\bar{p}(0)$ and $\bar{p}(t_f)$ respectively, within a finite amount of time $t_f$. We select a quintic function to represent this trajectory in continuous time. A sampling rate $\Delta t$ is selected and we sample the end-effector's trajectory to get a total of $N = \frac{t_f}{\Delta t}$ in-between poses. We proceed to evaluate the effective mass of the robot holding the object for each of these $N$ poses }
\label{Fig:Path}
\end{figure}
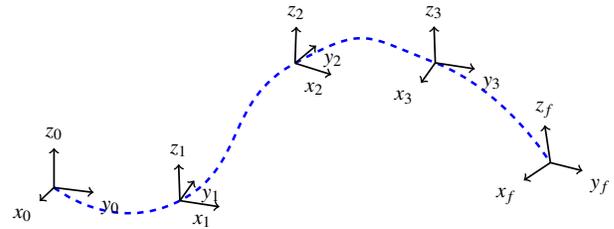
Let ${\mathrm{x}}(t) \in \mathrm{SE}(3)$ be a pose of end-effector at time $t$ where $0 \leq t \leq t_f]$, $\mathrm{SE}(3) = \mathbb{R}^3 \times \mathrm{SO}(3)$ and $\mathrm{SO}(3)$ denotes the group of rotations in three dimensions:  
$$\mathrm{SO}(3) = \{\mathrm{R} \in \mathbb{R}^{3\times 3}\; : \mathrm{R}\mathrm{R}^T = \mathrm{I},\; det(\mathrm{R}) = +1 \}.$$
We assume that an initial and a target pose of the object to be manipulated are recognised by a computer vision algorithm. Consequently, the poses of the robot's end-effector corresponding to those of the object are computed. As in~\cite{ramos2013time}, a polynomial function is used to generate a trajectory of end-effector poses necessary to move the object from the initial to the target pose.  For example consider the pick-and-place task of the book shown in Fig.~\ref{fig:RealExp} which is desired to be placed right next to the bottle. We represent the initial and final poses with $\mathrm{x}(0)$ and $\mathrm{x}(t_f)$ respectively where a quintic polynomial is used to generate the end-effector's trajectory. This polynomial has enough free parameters to ensure that the boundary conditions of position, velocity and acceleration are satisfied while the trajectory and its first and second order derivatives are continuous. An example of a trajectory is shown in Fig.~\ref{Fig:Path}.
\begin{align}\label{polynomial}
\begin{split}
\MoveEqLeft \mathrm{x}(t) = \mathrm{x}_0 + \mathrm{x}_1t + \mathrm{x}_2t^2 + \mathrm{x}_3t^3 + \mathrm{x}_4t^4 + \mathrm{x}_5t^5 \\ 
\MoveEqLeft \mathrm{x}(0) = x_0,~\mathrm{x}(t_f) = \mathrm{x}_{t_f} \\
\MoveEqLeft {\dot{\mathrm{x}}}(0) = {\dot{\mathrm{x}}}(t_f) ={\ddot{\mathrm{x}}}(0) = {\ddot{\mathrm{x}}}(t_f) = 0 
\end{split}
\end{align}
As mentioned in the introduction we need to compute the impact force applied by the manipulator during potential collisions, while it is manipulating an object with known inertial properties, namely mass, centre of mass (CoM) and inertia tensor. In case of simulated data or plain real-world shapes, we are able to calculate these properties easily. In the case of real-world manipulation of objects for which these properties are unknown, they can be estimated by using a variety of methods from the literature, e.g. \cite{Atkeson1985}\cite{Mirtich1996}\cite{Yu2005}.

We begin by presenting the augmented object model, as proposed in~\cite{Khatib1987}.    The \textit{augmented object model} describes the dynamics of a manipulator while holding an object. In our previous work~\cite{Mavrakis2016}, we used this augmented model to select grasps which minimised the post-grasp torque effort.

Consider the dynamic model of a robotic manipulator in the operational space expressed in Eq.~\eqref{RobotDynamicsEE}. While the analytical dynamics may not be known exactly in real applications, it is reasonable to assume that an approximation of the dynamic model can be identified.
\begin{equation}\label{RobotDynamicsEE}
\Lambda(\mathrm{x})\ddot{\mathrm{x}}+\mu(\mathrm{x},\dot{\mathrm{x}})+p(\mathrm{x}) = \mathrm{F}
\end{equation}
where $\Lambda(\mathrm{x}), \mu(\mathrm{x},\dot{\mathrm{x}})$ and $p(\mathrm{x})$ are the centrifugal, Coriolis and gravity force vector, respectively, acting in the operational space. $\mathrm{F}$ is the generalised force in the operation space including external wrench of force and torques applied to the manipulator from the environment.

While one can compute the force at every point of interest of the manipulator by writing the corresponding operational space equation, we can analyse the kinetic energy matrix $\Lambda(\mathrm{x})$ and compute the impact force during a collision without needing to solve the second order differential equation in Eq.~\eqref{RobotDynamicsEE}. 
However, this only represents the effect of the manipulator's dynamics on the impact force. According to previous studies in human-robot collaboration and safety, the impact force, as perceived by a human during collision with a manipulator, can be represented by the ``effective mass'', defined in \cite{khatib1995} as:
\begin{equation}\label{EffMass}
M = \frac{1}{v^T\Lambda^{-1}_{u}(\mathrm{x})v}
\end{equation}
where $v$ is a unit vector in the direction of motion and $\Lambda_{u}$ is the kinetic energy corresponding with translation only.

We are interested to quantify the impact force of the manipulator while it is moving an object with known inertial properties. Hence, we need to use the augmented object model. Let us consider the kinetic energy matrix of the object, in the object's CoM coordinate frame, denoted by $\pazocal{F}_{com}$, given by:
\begin{equation}\label{InerTens}
\Lambda_{O_{com}} = 
\begin{pmatrix}
m\mathrm{I}_{3x3} & {0} \\
{0} & I_{CoM}\\
\end{pmatrix} 
\end{equation}
where $m$ and $I_{CoM}$ are the object's mass and inertia tensor and $\mathrm{I}$ is the unit matrix.
% This kinetic energy can be also expressed in the frame attached to the end-effector at a desired grasping pose, denoted by~$\pazocal{F}_{gp}$ by using transformation $E$.
This kinetic energy can be also expressed in the frame attached to the end-effector at a desired grasping pose, denoted by~$\pazocal{F}_{gp}$ by using a transformation $T$, which transforms linear and angular velocities from ~$\pazocal{F}_{com}$ to ~$\pazocal{F}_{gp}$. Let $r$ be the vector that connects ~$\pazocal{F}_{gp}$ to ~$\pazocal{F}_{com}$ and $\hat{r}$ the cross-product operator for $r$, then $T$ is given by:
\begin{equation}
T = \begin{pmatrix}
I & \hat{r} \\
{0} & I\\
\end{pmatrix} 
\end{equation}
The kinetic energy matrix $\Lambda_{O_{gp}}$ expressed in ~$\pazocal{F}_{gp}$ is thus given by:
\begin{equation}\label{TransformT}
\Lambda_{O_{gp}} = T^T\Lambda_{O_{com}}T
\end{equation}
$\Lambda_{O_{gp}}$ needs to be transformed to an appropriate operational space coordinate representation. This is achieved by using a matrix ${E(\mathrm{x})}$ which relates operational generalised velocities to linear and angular velocities notation. ${E(\mathrm{x})}$ is only dependent on the choice of the variables to represent position and orientation in ~$\pazocal{F}_{gp}$ e.g, Cartesian position and Euler angles.  \cite{khatib1995}. The kinetic energy matrix of the object, expressed in the end-effector frames and in operational coordinates representation is given by: 
\begin{equation}\label{InerTenEE}
\Lambda_{obj}(\mathrm{x})= E^{T}(\mathrm{x})\Lambda_{O_{gp}}  E^{-T}(\mathrm{x})
\end{equation}
%%%%%%%%%%%%%%%%%%
%
% The following formula must be checked
%
%%%%%%%%%%%%%%%%%
% \todo{The following equation must be checked!!!!}
As the kinetic energy matrices $\Lambda_{obj}(\mathrm{x})$ and $\Lambda(\mathrm{x})$ are now represented in the same coordinate frame and operational coordinate representation, the kinetic energy of the whole system results from an addition of the two matrices:
\begin{equation}\label{TotalDynamicsEE}
\Lambda_{tot}(\mathrm{x}) = \Lambda_{obj}(\mathrm{x})+\Lambda(\mathrm{x})
\end{equation}
A more detailed explanation of the augmented model can be found in~\cite{Khatib1987}. Eq.~\eqref{TotalDynamicsEE} shows that the total kinetic energy matrix is simply the addition of the manipulator and the object energies expressed in $\pazocal{F}_{gp}$.  As $\Lambda_{tot}(\mathrm{x})$ has the same form and meaning as for the manipulator, namely $\Lambda(\mathrm{x})$, its inverse always exists and it has the following form:
\begin{equation}\label{LamdaInv}
\Lambda_{tot}(\mathrm{x})^{-1} = \begin{pmatrix} \Lambda^{-1}_{u_{tot}}(\mathrm{x}) & \Lambda_{uw_{tot}}(\mathrm{x}) \\
\Lambda^{T}_{uw_{tot}}(\mathrm{x}) & \Lambda^{-1}_{w_{tot}}(\mathrm{x})
\end{pmatrix}
\end{equation}
where $\Lambda_{u_{tot}}(\mathrm{x})$ represents the inertial properties of the augmented model associated with translation, $\Lambda_{w_{tot}}(\mathrm{x})$ is the inertial properties associated with rotation, and $\Lambda_{uw_{tot}}(\mathrm{x})$ represents a measure of coupling between angular and linear parts.

As mentioned in the introduction, we aim at selecting a grasp that produces minimum impact force in case of a collision, as compared to other possible grasps. We assume that a set of possible grasps for the object have been generated by a grasp synthesizer, e.g. \cite{kopicki2015one} or several other recently proposed methods. Since the object's CoM is known, every synthesized grasp represents a pose difference between the CoM frame and the robot's end-effector frame.

It has been shown that a manipulator during a collision is perceived according to its effective mass (Eq.~\eqref{EffMass}).
In analogy, we define the effective mass of the total system as
\begin{equation}\label{EffMassTot}
M_{tot} = \frac{1}{v^T\Lambda^{-1}_{u_{tot}}(\mathrm{x})v}
\end{equation}

The total effective mass is dependent on the object's inertia tensor w.r.t $\pazocal{F}_{gp}$. This means grasping from different poses results in different kinetic energy values of the corresponding augmented models. Hence, there may exist one grasp choice which could result in an impact force lower than that for all other possible grasps. In the next section we show that grasp poses have a significant effect on a safety measure of effective mass in the context of human-robot collaboration. 
%%%%%%%%%%%%%%%%%%%%%%%%%%%%%%%%%%%%%%%%%%%%%%%%%%%%%%%%%%%%%%%%%%%%%%%%%%%%%%%%%%%%%%%%%%%%%%%%%
%
%
%
% Amir Checked PROBLEM FORMULATION SECTION in the above : 010320170336
%
%
%
%%%%%%%%%%%%%%%%%%%%%%%%%%%%%%%%%%%%%%%%%%%%%%%%%%%%%%%%%%%%%%%%%%%%%%%%%%%%%%%%%%%%%%%%%%%%%%%%%

\iffalse
\subsubsection{Null Space Task Decomposition}
In the case of a redundant manipulator, the relationship between joint velocities $\dot{q}$ and end-effector velocities $\dot{p}$ can be formulated as
\begin{equation}\label{Redundant}
\dot{q} = J^{\dagger}\dot{p}+(I-J^{\dagger}J)k
\end{equation}
where $J$ is the manipulator's Jacobian, $J^{\dagger}$ is the pseudo-inverse of the Jacobian, $I$ is the unit matrix and $k$ is a vector. The vector $k$ acts on the Jacobian null space, and as a result, does not have an effect on the end-effector movement. Usually, the vector $k$ is chosen so as to minimize a selected function, enabling the manipulator to execute this secondary task in parallel with the end-effector movement. The vector $k$ is given by
\begin{equation}\label{vectorK}
k = -\frac{\partial U(q)}{\partial q}
\end{equation}
where $U(q)$ is the function to be minimized. The negative sign indicates minimization. We select the following as a cost function to be minimised:
\begin{equation}\label{minFunction}
U(q) = M^{*} = \frac{1}{u^T\Lambda^{-1 *}_{u_{tot}}(\mathrm{x})u}
\end{equation}
where the star symbol indicates the optimal effective mass found in the previous step. 
\fi

\section{Experimental results}
This section presents the results of several experiments including two experiments in simulation environments and one experiment using a real robot. The experiments are designed such that the alteration of the robot's dynamics with each grasping point is illustrated. In particular, pre-calculated effective mass along a given trajectory for each grasp, and the impact force that the grasp will produce in a collision scenario, are discussed in the following to show the effectiveness of our approach.

We use the 7-DOF Baxter{\textregistered} robot by \textit{Rethink Robotics} for both simulation and real robot experiments. 
We use the Baxter Software Development Kit (SDK), the Python Kinematic Dynamic Library (PyKDL) and the Gazebo simulation environment which supports object and sensor generation and built-in physics engines\footnote{Orocos Kinematics and Dynamics Library can be found at: http://www.orocos.org/kdl. The Gazebo simulation environment can be found at: http://gazebosim.org/.}. 

We sample the continuous end-effector trajectory with a sampling rate of $\Delta t$, to collect a total number of $N = \frac{t_f}{\Delta t}$ intermediate end-effector poses $\mathrm{x}_i ,~i = 1,. . .,N$ in Cartesian space. The poses $\mathrm{x}_i$ are used to calculate the dynamic properties of the manipulator during the trajectory.

% \subsection{Simulation}

\begin{figure}[t]
\centering
\begin{subfigure}[b]{0.63\linewidth}
\includegraphics[width = 1\linewidth]{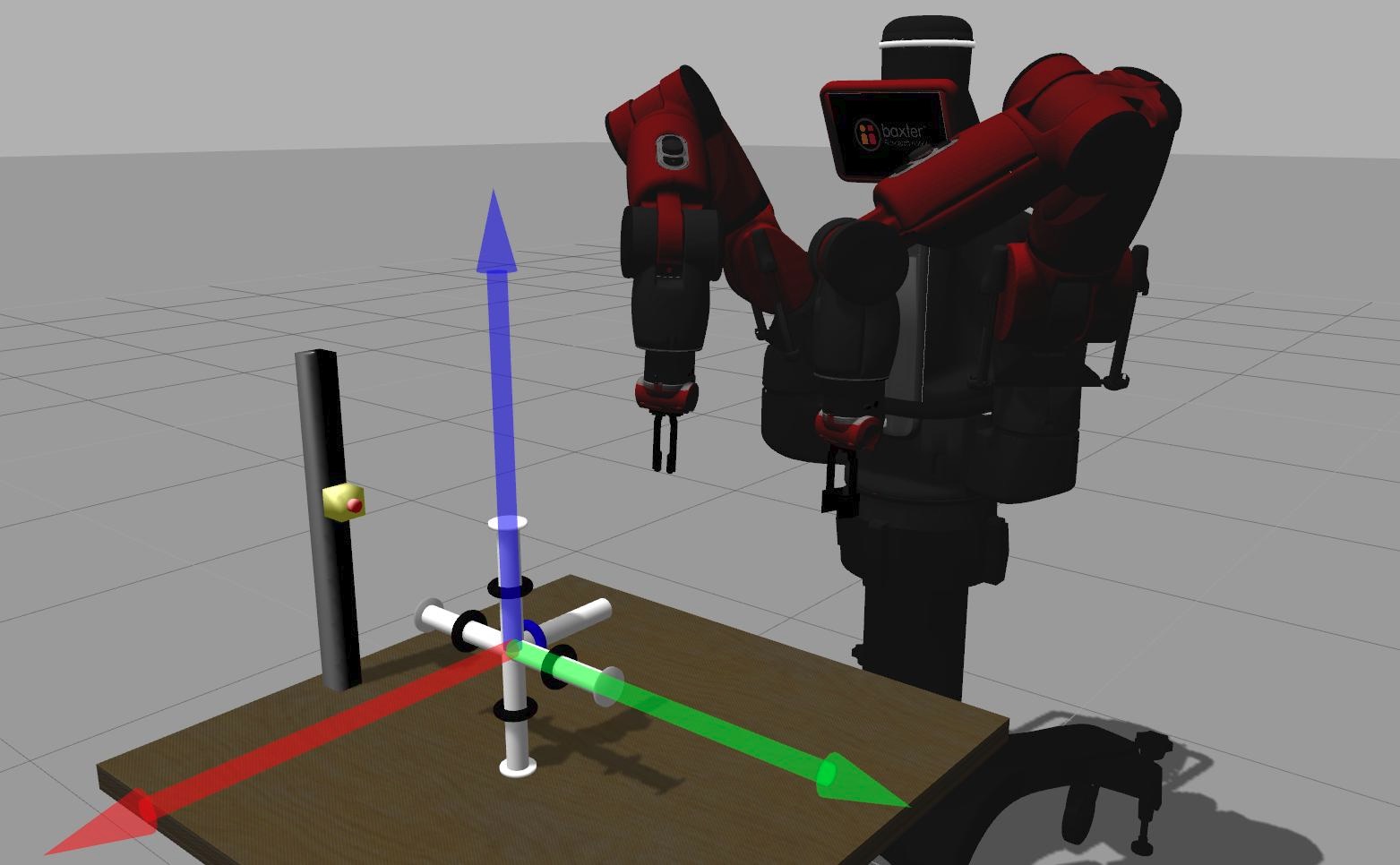}
\caption{\label{Fig:Baxter}}
\end{subfigure} \vspace{0.1pt}
\begin{subfigure}[b]{0.3  \linewidth}
    \includegraphics 
    [width = 1\linewidth]{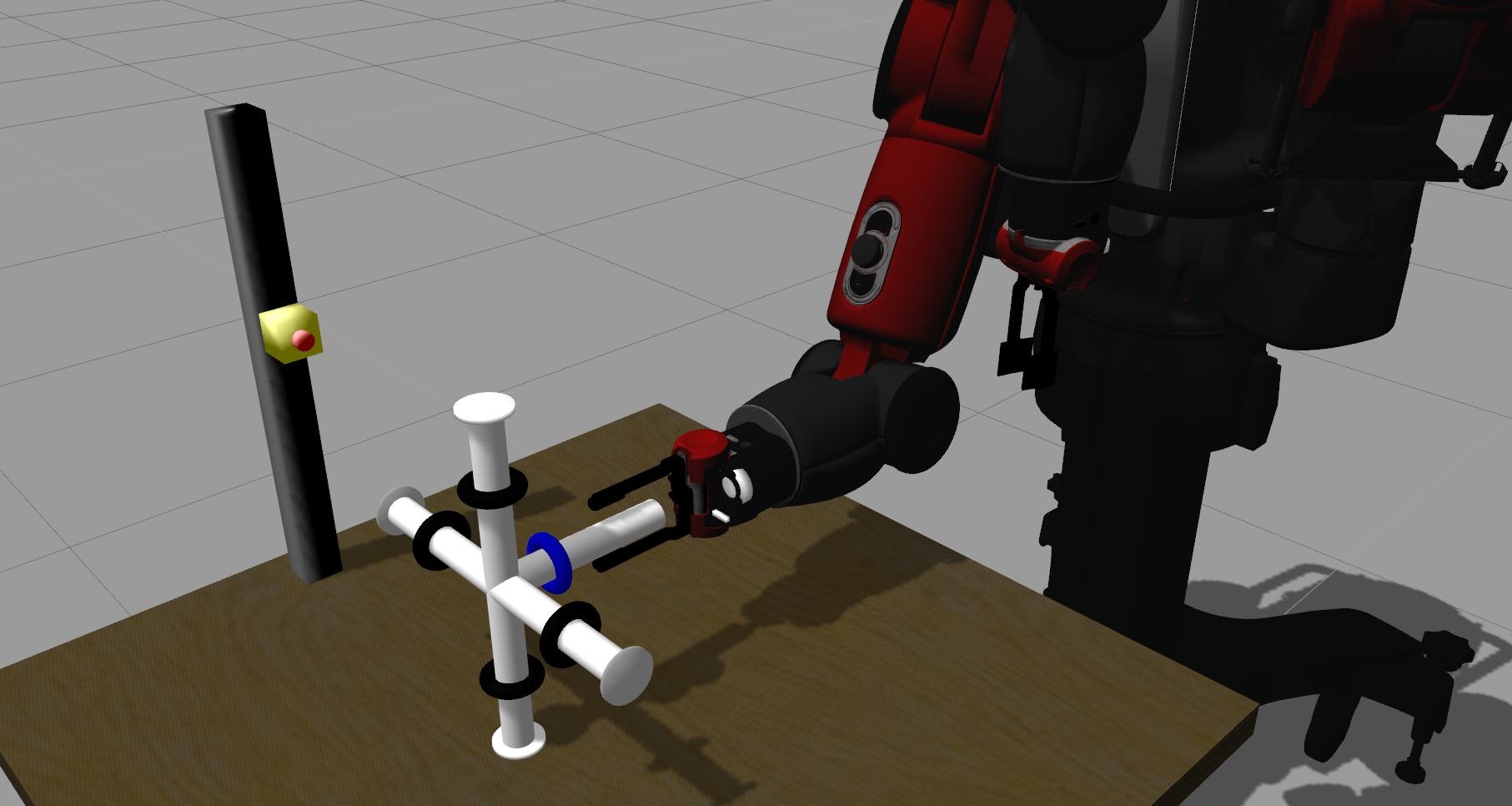}
   \caption{\label{Fig:StartingPoint}}

    \includegraphics[width = 1\linewidth]{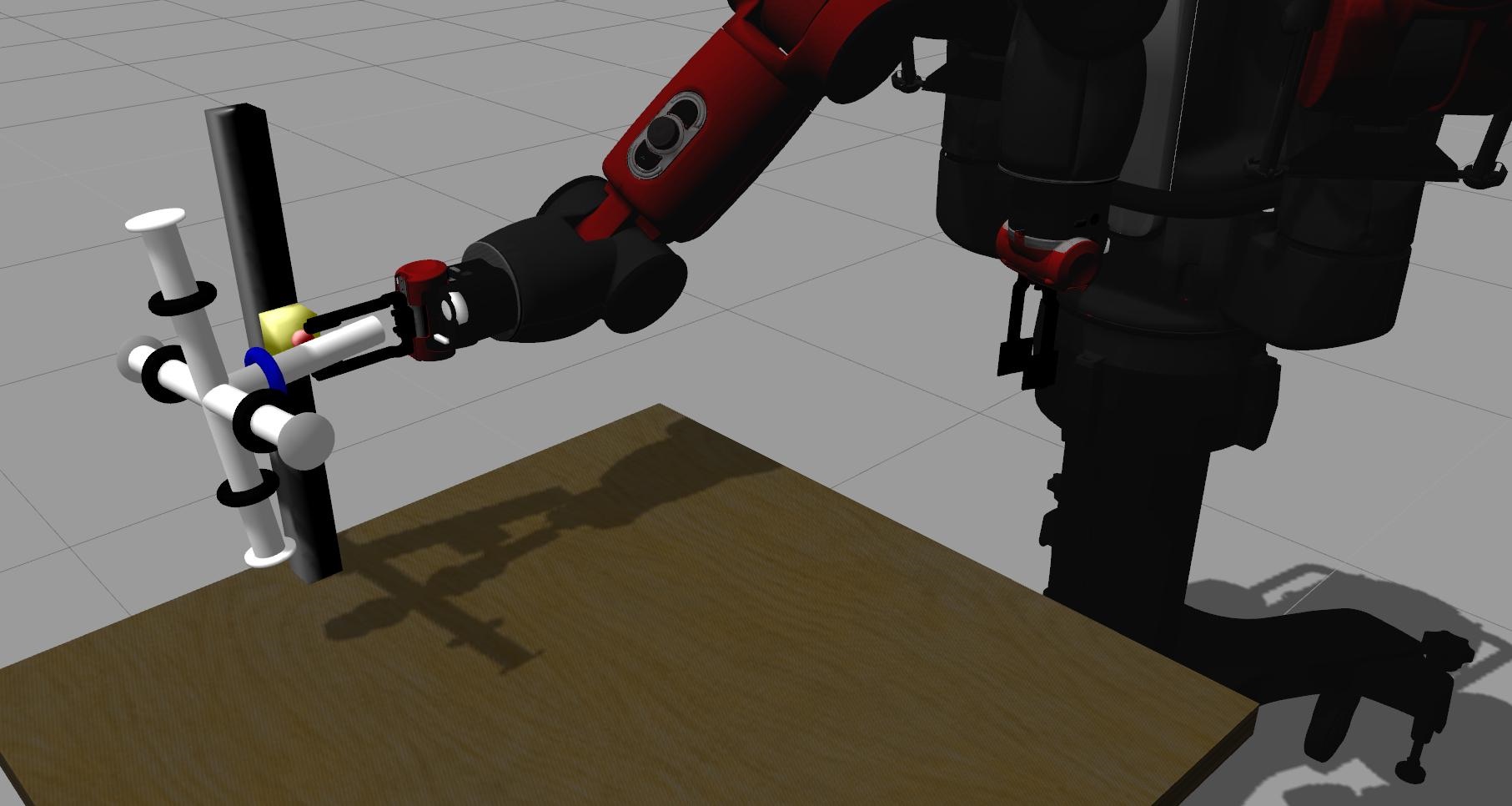}
    \caption{\label{Fig:EndingPoint}}
\end{subfigure}%
\caption{The Baxter robot and tensor object simulation configuration. We use an object with variable inertia tensor to demonstrate the effect of different grasping positions on the severity of collisions with the environment. Our experiments consist of the robot approaching the object, and grasping it as shown. The robot then lifts the object and transports it along the task trajectory. At some point along this trajectory, we intentionally introduce a rigid pillar equipped with a virtual force sensor, shown here as the red button. The robot end-effector then collides with the force sensor. Our purpose is to measure the exerted force on the sensor along the path direction.(\subref{Fig:Baxter}) The simulation configuration. (\subref{Fig:StartingPoint}) The starting point of the end-effector path. The grasping point $gp$ identifies with the starting point. (\subref{Fig:EndingPoint}) The collision point of the end-effector path.}
\label{Fig:TO}
\end{figure}

\subsubsection{Simulation with Tensor Object}
\begin{figure}[h]
 \centering
\includegraphics[scale = 0.1]{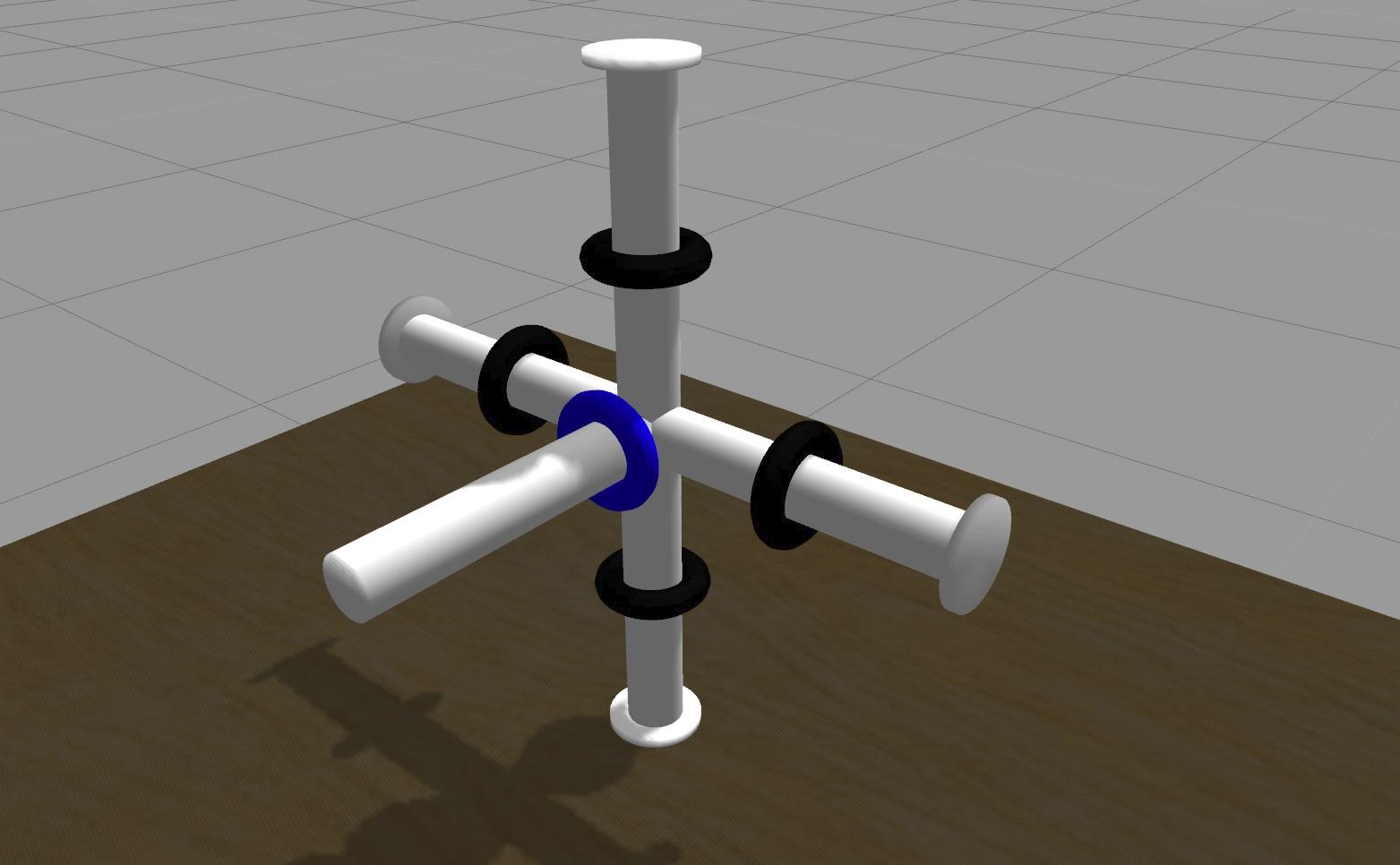} 
\caption{~The tensor object. The caps of the cylinders have negligible mass. The black and blue weighted rings are able to move along the cylinders and latch when needed. By changing a ring's position, we are able to change the inertia tensor of the object and the perceived inertia tensor at the contact point. The ring on the $x$ axis (handle) is coloured blue so that it can easily be distinguished.} \label{Fig:tensorObject}
\end{figure}

In the first simulation, we use a ``tensor object''. Tensor objects are widely used in experimental psychology, to demonstrate how changes in an object's inertia tensor affect its perceived properties by a human who is manipulating it~\cite{Amazeen1996}. A tensor object consists of five cylinders in the form of a 3D coordinate frame, as shown in Fig.~\ref{Fig:tensorObject}. One of the cylinders is chosen as the handle for grasping. Toroidal weights (shown as black and blue coloured rings in Fig.~\ref{Fig:tensorObject}) are located on each cylinder and can slide along it. The total mass of the tensor object is $0.43$ kg. %The robot holds the object from a pre-defined contact point $CP$ located on the handle. 
By changing each weight's position along its respective cylinder, the object's inertia tensor w.r.t. the CoM can be modified. Subsequently, the perceived inertia tensor at  $\pazocal{F}_{gp}$ is changed.

We have designed our tensor object, shown in Fig.~\ref{Fig:tensorObject}, such that the robot pose expressed by $\pazocal{F}_{gp}$ relative to the $\pazocal{F}_{com}$ is kept invariant over different grasps, whereas the inertia tensor of the object can be changed. We keep velocity and type of collision invariant over all experiments. 
% in both experiments to ensure that changes in impact forces are a product of the robot's augmented dynamics.
% Since we wish to show how perceived inertia tensors on the end-effector will affect the robot dynamics, we use a \textit{tensor object}. 
This allows us to study the effect of just one variable (inertia tensor of the object being manipulated) on the impact force during a post-grasp collision, while keeping all other conditions invariant throughout the experiment.  % As mentioned, by using a tensor object and changing the inertia tensor, we impose the equivalent of a different physical grasp on an object, without actually moving the manipulator to the physical grasp location. As will be shown below, the reason to use this object is to ensure that any change in the collision magnitude is attributed only to the change of the perceived inertia tensor and not on the change in the manipulator's posture.

\begin{figure}[]
\centering
\includegraphics[width=\linewidth]{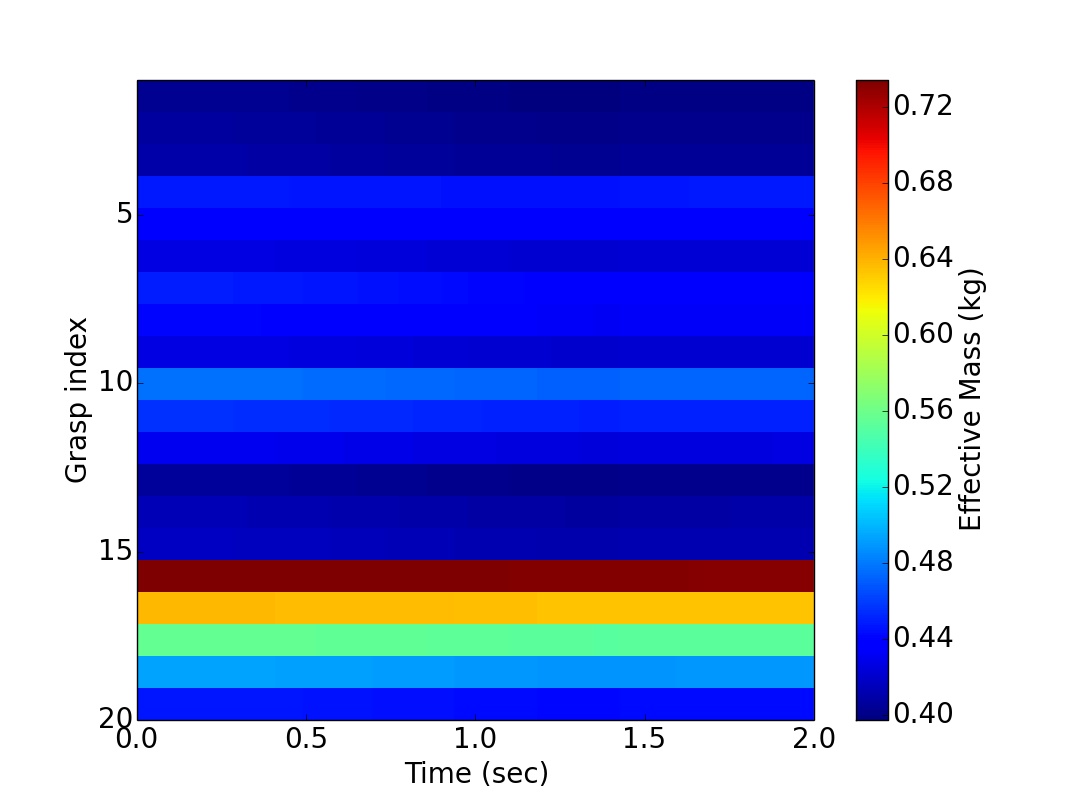}
\caption{Map of the effective mass along the trajectory for all grasps of the tensor object. The vertical axis denotes each of the 20 example grasps, and the horizontal axis denotes time, throughout the duration of the post-grasp trajectory. Colour denotes the magnitude of the effective mass for each grasp at each time step. Each horizontal segment, starting from left to right, plots the magnitude of the effective mass that each grasp produces over the entire time duration of the task trajectory. It can be seen that different grasps can indeed generate different effective masses along the duration of the task trajectory, with significant variations in the effective mass magnitude. The result of these variations is that our methodology can discriminate safer from less safe grasps.    }  
\label{Fig:allGraspsImage}
\end{figure}

$20$ different grasps were generated by altering the position of the rings on the tensor object. For each grasp
we prepared a collision scenario to measure the impact force that is delivered by the robot end-effector. % More specifically, we generated $20$ different grasps by varying the position of the rings on the tensor object. 
The robot picks and moves the object to its right side according to the task trajectory described in the previous section. The initial and final 3D positions of the end-effector were set to $p_0 = (1,~0,~0.03)$ and $p_f = (1.1,~-0.38,~0.16)$ respectively. We kept the orientation unchanged during the trajectory. The trajectory total duration is 2 seconds. After a brief time, the end-effector collides with the force sensor on the vertical pillar (Fig.~\ref{Fig:TO} ).
% A task was given to the robot in the same form as described in Section IIIa. 
%The initial and final 3D positions of the end-effector were set to $p_0 = (1,~0,~0.03)$ and $p_f = (1.1,~-0.38,~0.16)$ respectively. We kept the orientation unchanged during the task execution to $q = (0.5,~0.5,~0.5,~0.5)$ using quaternion notation. 
%The task duration was set to $2$ sec. 
We use the approach proposed in the previous section to pre-calculate the effective mass of the augmented model corresponding to each grasp, and the results are shown in Fig.~\ref{Fig:allGraspsImage}.
The results show that the value of effective mass significantly varies between different grasps. This figure indicates the safest grasp with the minimum effective mass.

To further validate our hypothesis, that grasps with different effective mass produce significantly different impact forces, we simulated a collision while recording the impact forces for various different grasps. %A switch with a force sensor is placed at a constant point of the predetermined path. The sensor measures Cartesian force w.r.t. world frame. Since we know the end-effector geometric trajectory and direction of motion $v$, we can project the measured 3D force in this direction.

\begin{figure}[]
\centering
    \includegraphics[width = 1\linewidth]{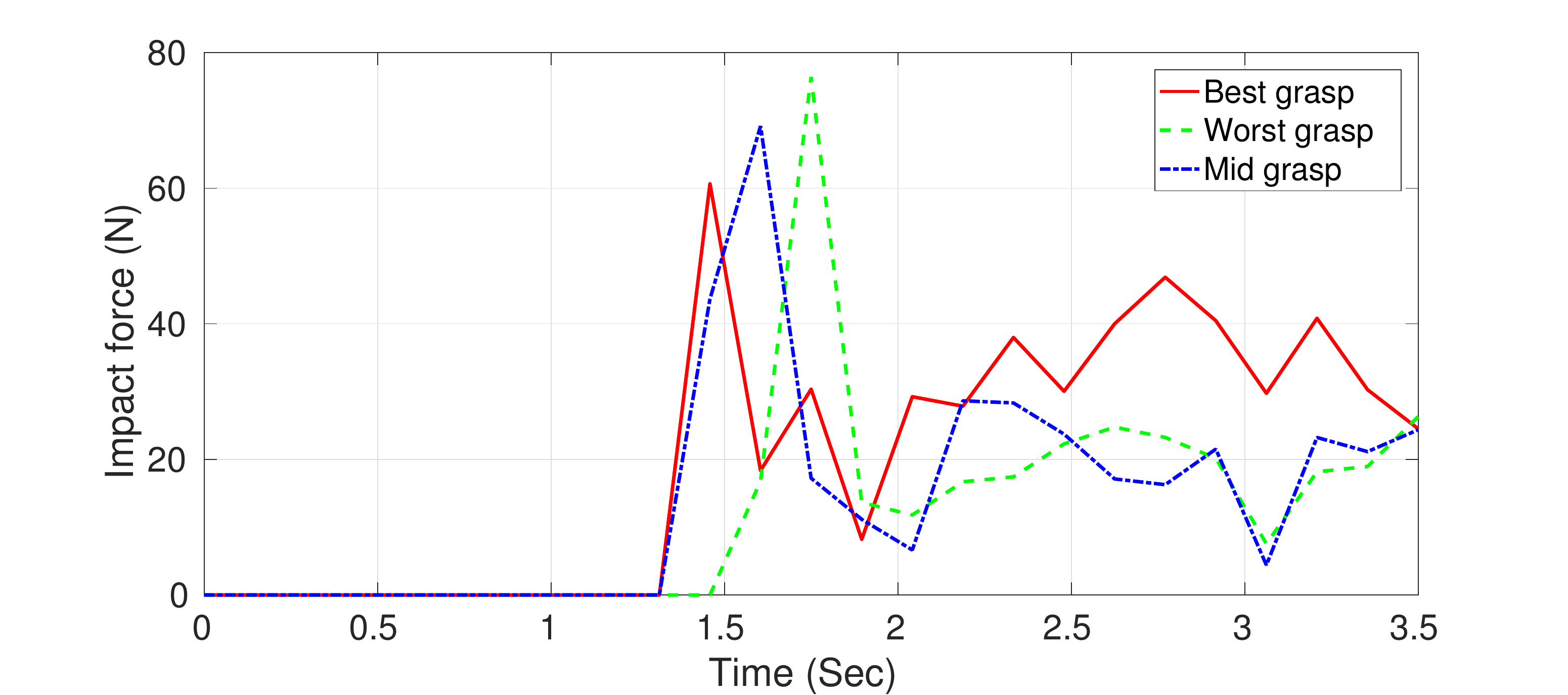}

\caption{Impact force profiles for three different grasps of the tensor object. Red colour corresponds to the minimum effective mass grasp, green to the maximum effective mass grasp, and blue to a grasp with an intermediary value, sampled from the $20$ grasps shown in Fig.\ref{Fig:allGraspsImage}. The impact forces are visible as the large initial force peak, and it is clear that the magnitudes of these impact forces vary between the grasps. Since the task trajectory is the same for all grasps and the collision type and geometry remain the same, these variations in the impact forces depend only on the choice of grasp. The relatively high magnitudes are attributed to the fact that the robot collides with a simulated, rigid pillar that has a virtual force sensor. As the robot keeps pushing, the recoil generates oscillations (especially in the spring-loaded series-elastic-actuators of the Baxter robot) which eventually converge to a steady state contact force.}  
\label{fig:Collision}
\end{figure}

%When a collision occurs the impact force is affected by the local geometry of collision, but in all cases the maximum force is applied in the direction of motion. Thus, we provide a measure of the maximum possible force to be exerted in the collision. We program the robot to grasp the object and follow the task, only to produce a collision. 
Fig.~\ref{fig:Collision} plots the impact forces resulting from three different grasps: the grasp with maximum effective mass in Fig.~\ref{Fig:allGraspsImage}; the grasp with minimum effective mass; and a grasp with moderate effective mass. The robot is commanded to cease its effort 2 seconds after the collision, so that the peak impact force and the steady state contact force values can be easily evaluated. Because the collision happens roughly 1.5 seconds after the start of the trajectory, the total measurement time was 3 seconds as shown in Fig.~\ref{fig:Collision} . The impact forces are clearly visible in Fig.~\ref{fig:Collision} as the large peak in the force signals at the collision time. It can be seen that the values of these peak forces are in accordance with the computed effective mass, i.e. the greater the effective mass, the greater the peak value of the impact force. 
These measurements validate our hypothesis (Fig.~\ref{Fig:allGraspsImage}) and suggest that our proposed approach can make a useful contribution to improving robot safety.
%The point of minimum effective mass results in lower force during a collision, the point of maximum effective mass in higher force, and the point in between results in moderate force. 
%%%%%%%%%%%%%%%%%%%%%%%%%%%%%
%
%CEHCKED UP TO THIS POINT 010320170813
%
%
%%%%%%%%%%%%%%%%%%%%%%%%%%%%%

\subsubsection{Simulated grasping and moving of book object}
\begin{figure}[]

\begin{subfigure}[]{1\linewidth}
\centering
 \includegraphics[width = 0.7\linewidth]{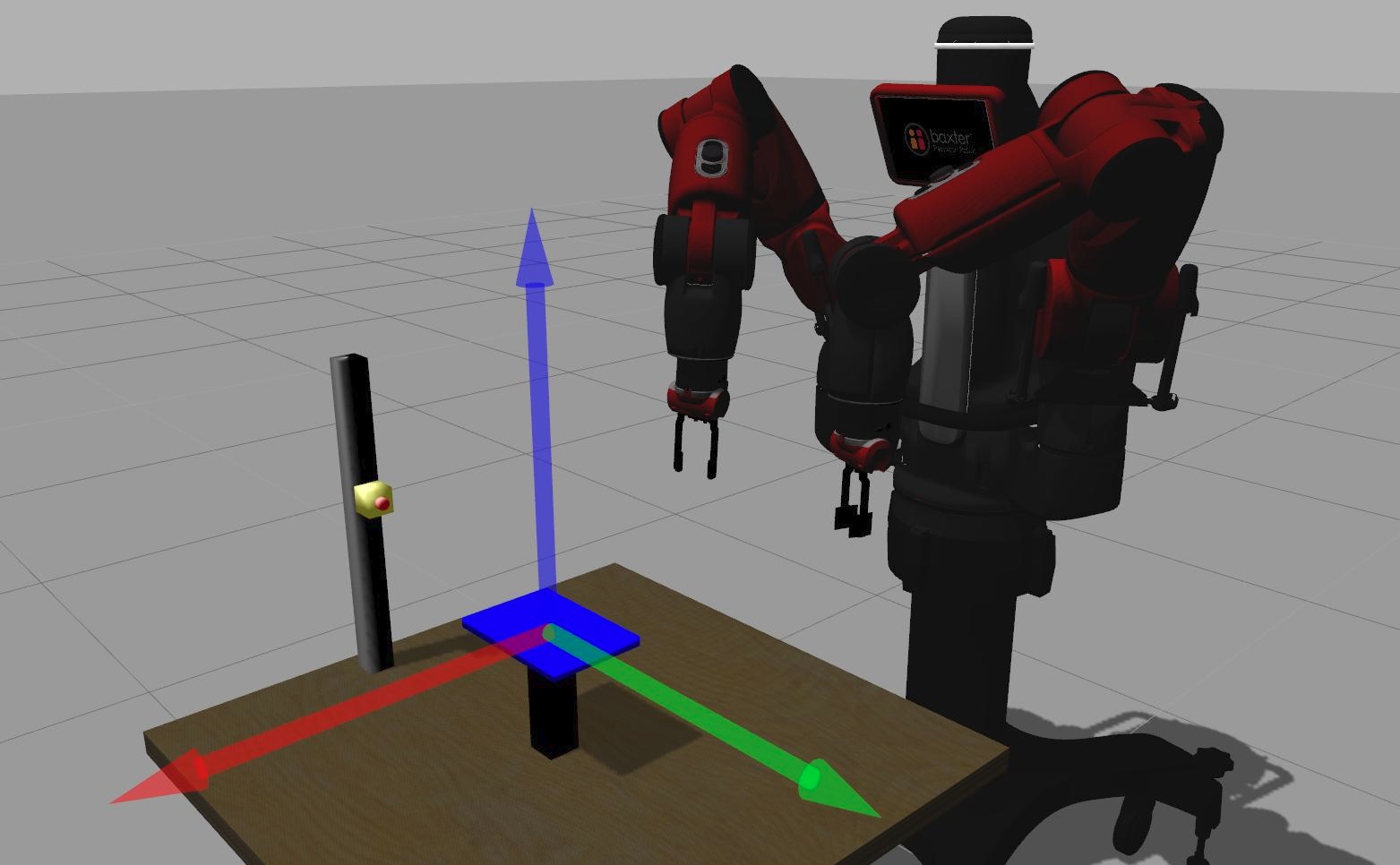}
   \caption{ \label{Fig:BookConfig}}
\end{subfigure} 
\centering
\begin{subfigure}[]{0.25\linewidth}
 \includegraphics[width = 1\linewidth]{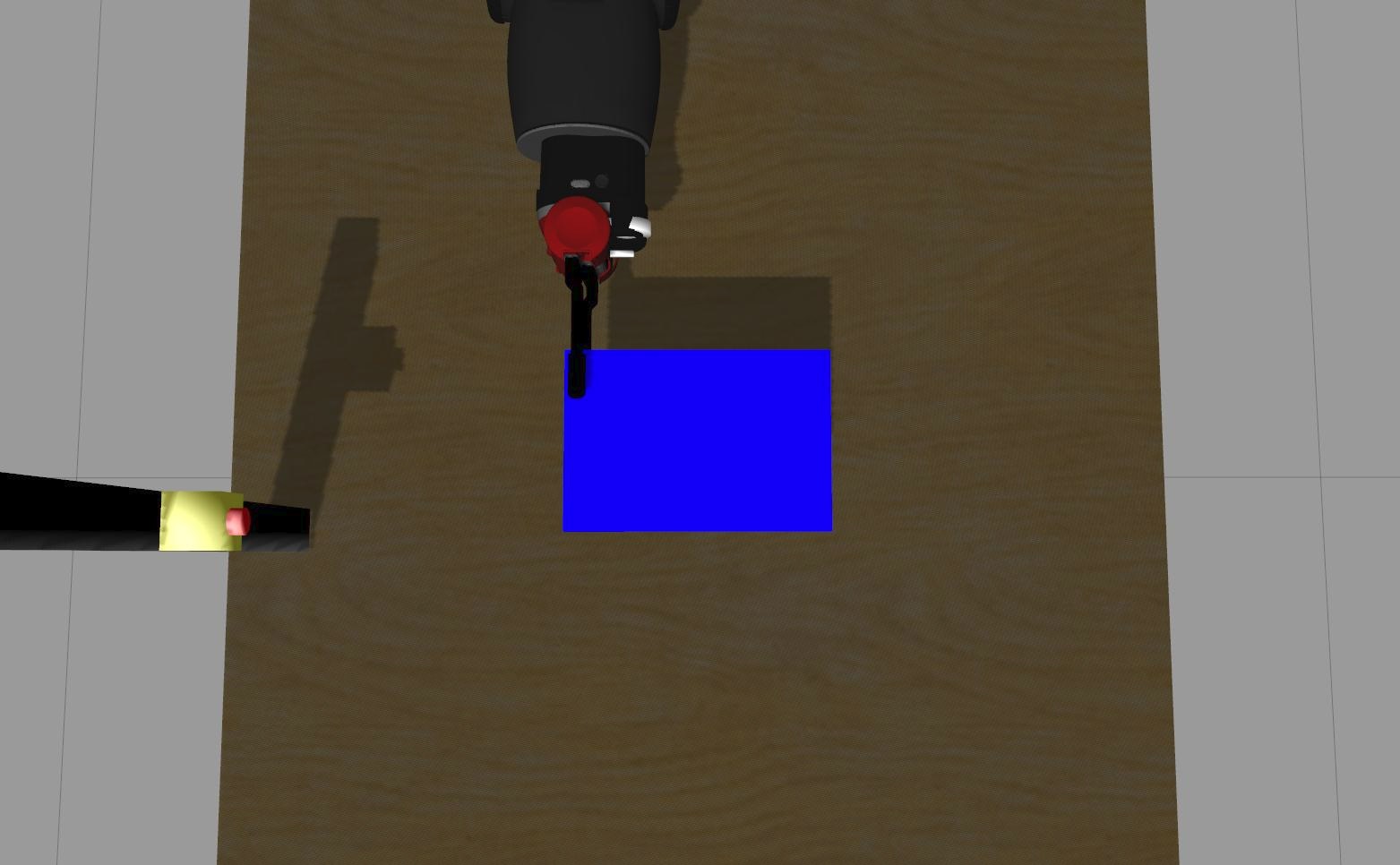}
  \caption{  \label{Fig:BookGrasp1}}
  \end{subfigure}
  \begin{subfigure}[]{0.25\linewidth}
\includegraphics[width = 1\linewidth]{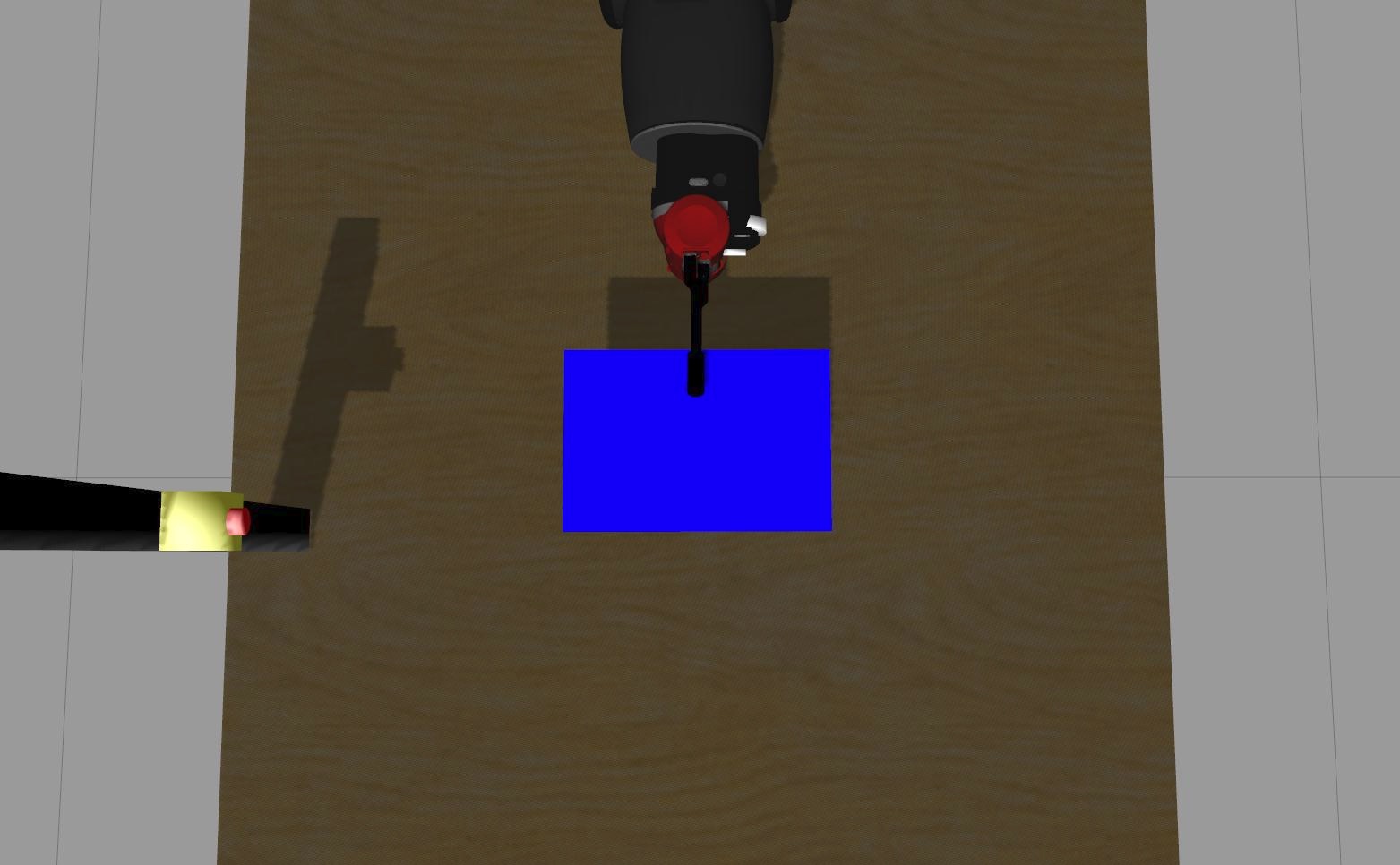}
  \caption{  \label{Fig:BookGrasp2} }
  \end{subfigure}
 \begin{subfigure}[]{0.25\linewidth}
 \includegraphics[width = 1\linewidth]{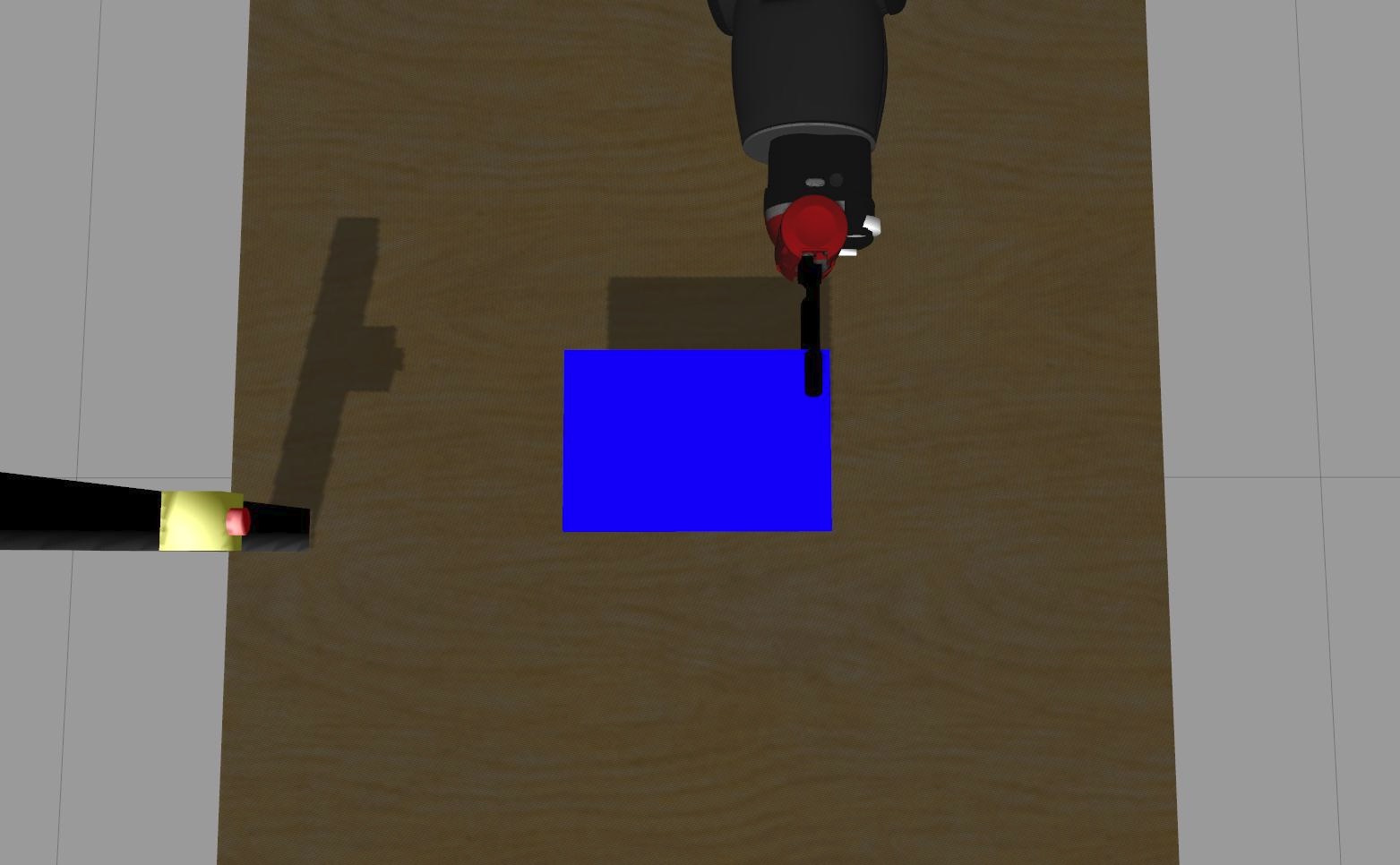}
    \caption{\label{Fig:BookGrasp3}}
\end{subfigure}
\caption{Simulated grasping and moving of a book. The robot was given three grasping points on a book with dimensions $22$x$15$x$1.5$ cm and mass of $0.34$ kg. After grasping the book, the robot hits the force sensor shown on the black pillar, and the impact forces are measured. (\subref{Fig:BookConfig}) Experimental setup. The book coordinate frame is visible. (\subref{Fig:BookGrasp1}) First grasp. (\subref{Fig:BookGrasp2}) Second grasp. (\subref{Fig:BookGrasp3}) Third grasp.  }  
\label{fig:BookGrasps}
\end{figure}

\begin{figure}[t]
\centering
\includegraphics[width = 1\linewidth]{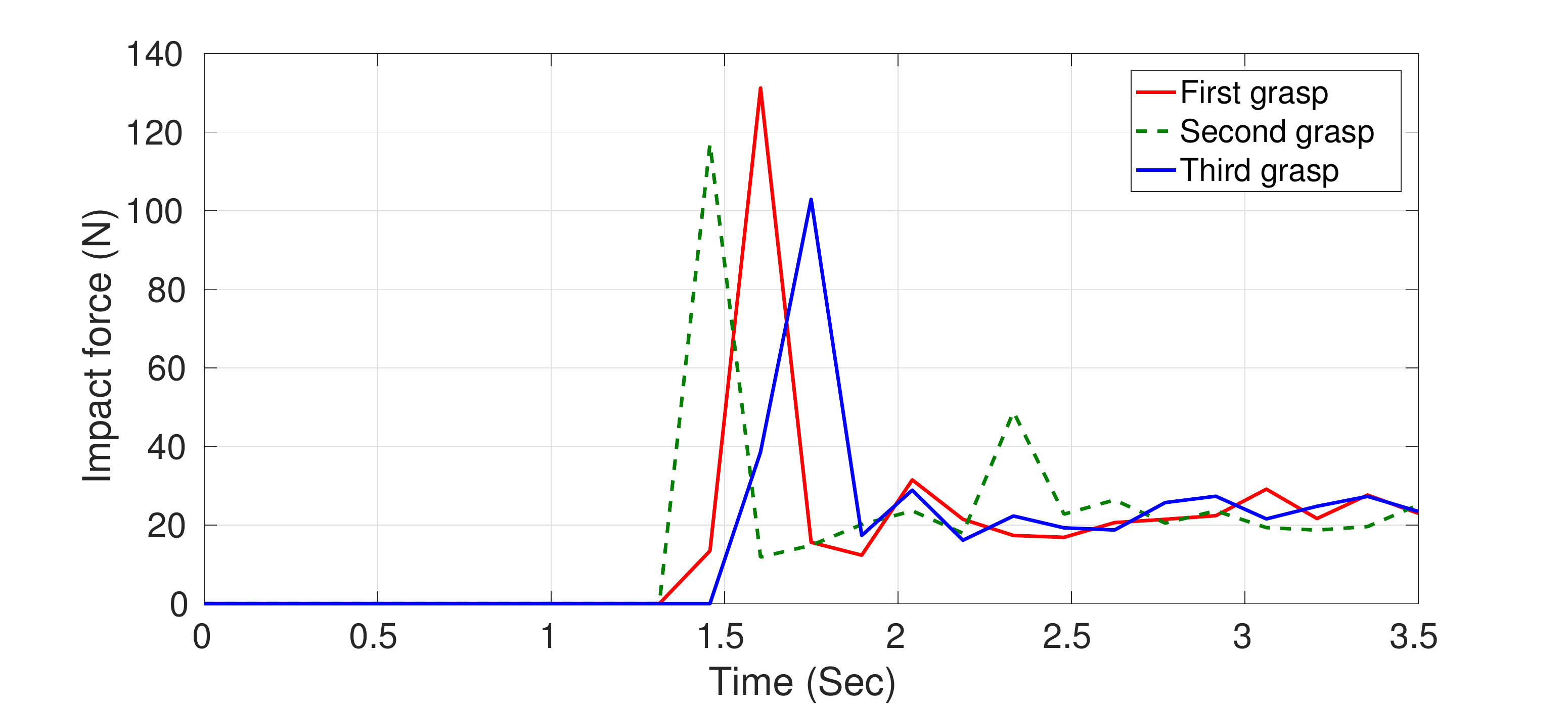}
\caption{Simulated impact force evolutions, between the robot's wrist and a virtual force sensor, for three different choices of grasp. Similar to the tensor object experiment, the three different grasps on the book object yield significantly different impact force values during the same collision profile. The forces for the first, second and third grasp are shown in red, green and blue respectively. The robot continues pushing after colliding, and the inherent elasticity of the Baxter actuators leads to oscillations which then decay to a steady state contact force. Note that the different timings of the initial impact in each case, are because the robot's wrist begins its motion at three different positions, corresponding to three different grasps on the spine of the book.}  
\label{fig:BookCollisionPlots}
\end{figure}

We next apply our approach to a more realistic example in which several different grasp locations on an object are attempted and, consequently, the robot's configurations and post-grasp trajectories become altered for each new grasp (in contrast to the previous experiment, in which the specially designed tensor object enabled identical robot motions over all experimental runs).  
%In the previous simulation we kept the manipulators posture constant and we changed the inertia tensor of the object. To see the performance of potential slight changes in the manipulator's posture when is grasping from different physical locations, while on the same time varying the inertia tensor of the object, 
We simulated a book-shaped object with dimensions $22$x$15$x$1.5$ cm and mass of $0.34$ kg (to align with our real robot experiments with a real book, described in the next section). This book object is initially positioned in front of the robot on a table. The robot performs three different grasps, on three different parts of the book, as shown in Fig.~\ref{fig:BookGrasps}. 
We use task setup similar to the previous experiment, i.e. after grasping the book, the robot collides forcefully with a rigid pillar containing a virtual force sensor.
The grasp points are at $-0.1$, $0$ and $0.1$, where these are grasp positions along the spine of the book, which is aligned with the y axis (shown in Fig.\ref{Fig:BookConfig} as the green arrow). The pre-calculated effective masses along the post-grasp trajectory, for each of the three grasps, are shown in Fig.~\ref{fig:BookGrasps}. For each grasp, we used the force sensor to model the resulting impact forces, which are plotted in Fig.~\ref{fig:BookCollisionPlots}.
Again, bigger impact forces result from grasps which yield bigger effective masses along the post-grasp trajectory, suggesting that computation of effective mass is  a useful predictor of collision safety. Furthermore, it is clear that significantly different effective mass, and significantly different impact forces, result from different choices of grasp.

% In all cases, the actual magnitude of the force can be calculated from the augmented dynamic model, the collision geometry, the type of collision and the velocity of the manipulator.

% Nevertheless, since in our case the collision point does not change, the effective mass in the instance of the collision remains unaffected. As a result, any difference in impact force measurements at the collision instance is attributed only to the relative position of the object w.r.t. the end effector.

\begin{figure}[t]
\centering
\includegraphics[width = 1\linewidth]{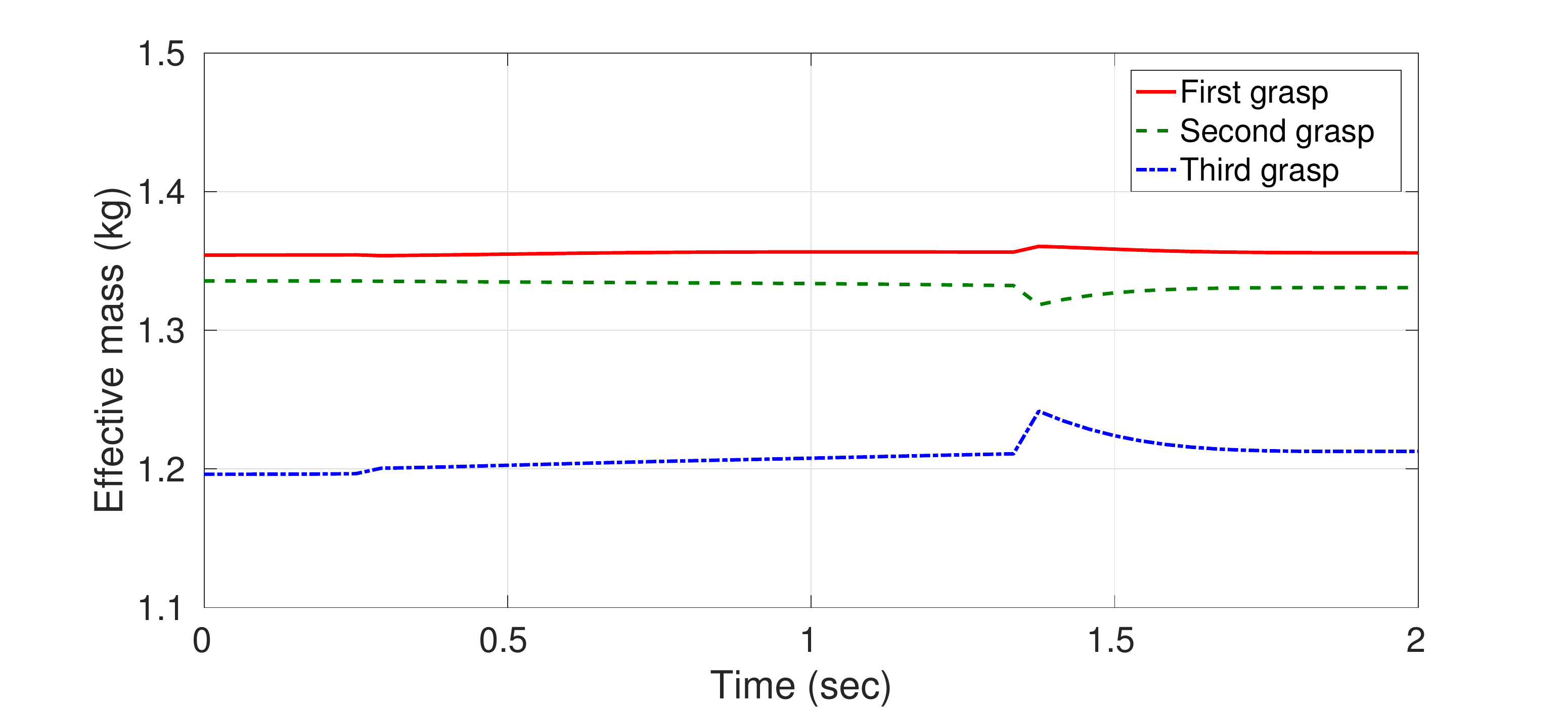}
\caption{Effective mass values, computed along a post-grasp trajectory, for three different grasping points on the book object. It can be seen that different grasps result in different effective masses. Computing effective mass along the desired post-grasp trajectory, can be used to predict the safety of each grasp w.r.t. collisions.  }  
\label{fig:BookGrasps}
\end{figure}

\subsubsection{Experiment using a real robot}
To further test our approach, we repeat the simulated book experiment using a real Baxter robot. We keep the experimental set-up consistent with that of the simulation experiment, i.e. identical grasp positions and post-grasp trajectories. The real book, shown in Fig.~\ref{fig:RealExp} has exactly the same mass and dimensions as the virtual book object in the simulation experiment, so that our effective mass computation, Fig.~\ref{fig:BookGrasps}, is identically valid for our experiment with the real Baxter robot and the real book. To avoid damaging expensive equipment, we make the real robot collide with a plastic bottle full of water, instead of a completely rigid pillar as used in the simulation experiment. During the post-grasp trajectory, the robot collides with the bottle for a brief period of time, knocking it over.

\begin{figure}[]
\centering
\begin{subfigure}[]{0.8\linewidth}
\centering
\includegraphics[width = 0.95\linewidth]{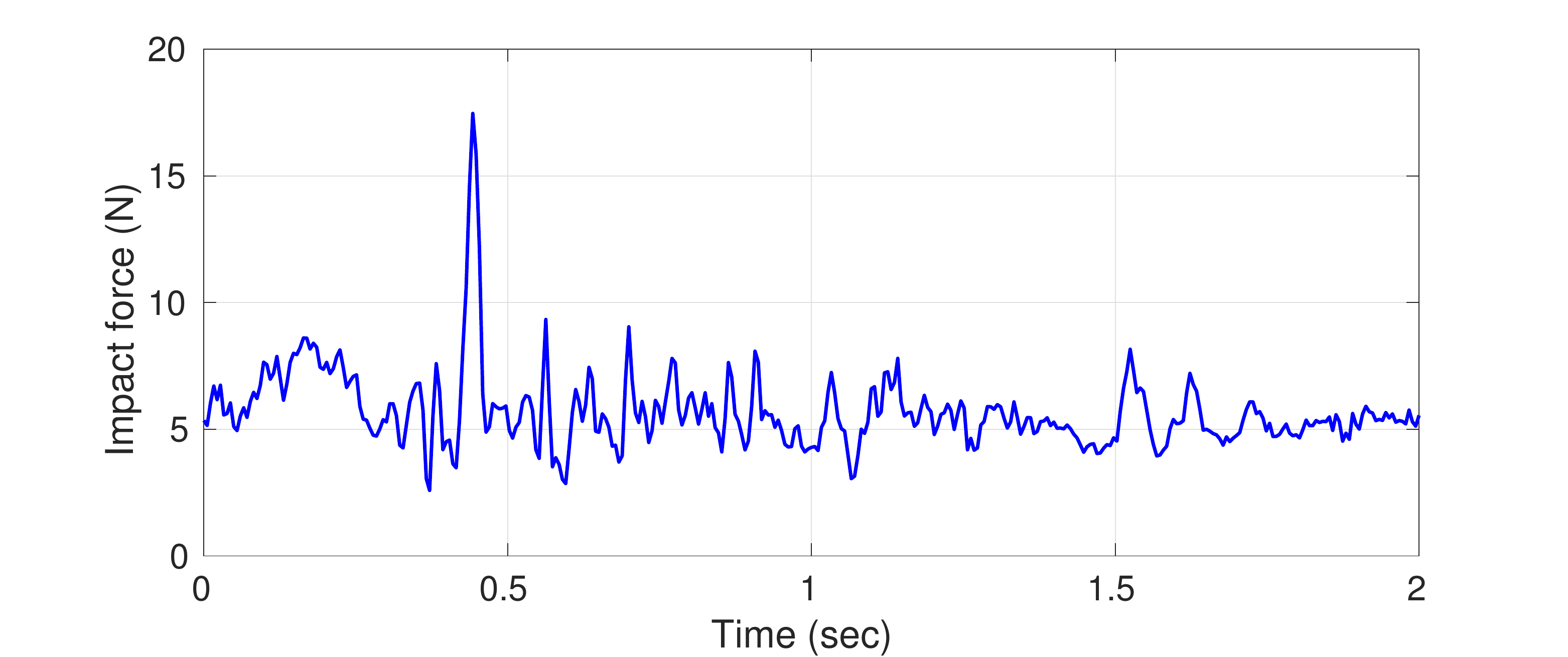}
\caption{\label{Fig:RealGrasp1} }
\includegraphics[width = 0.95\linewidth]{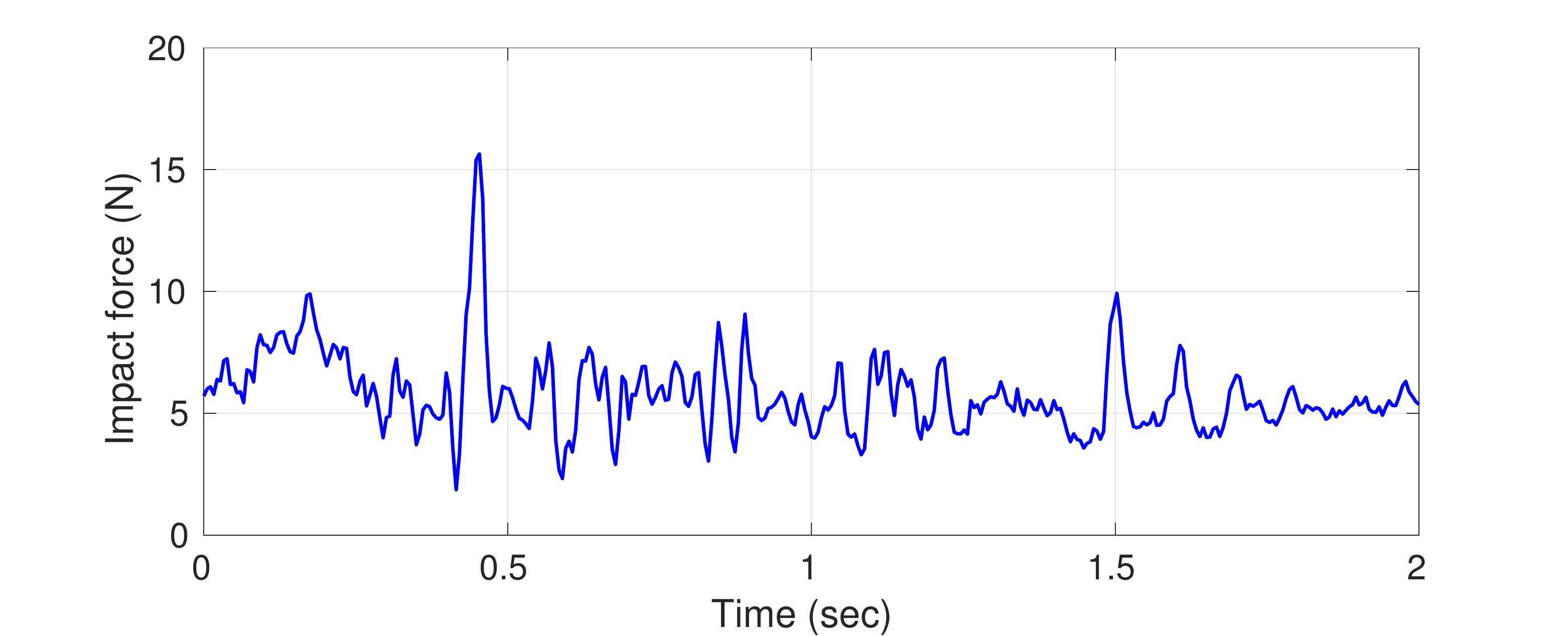}
\caption{\label{Fig:RealGrasp2} }
\includegraphics[width = 0.95\linewidth]{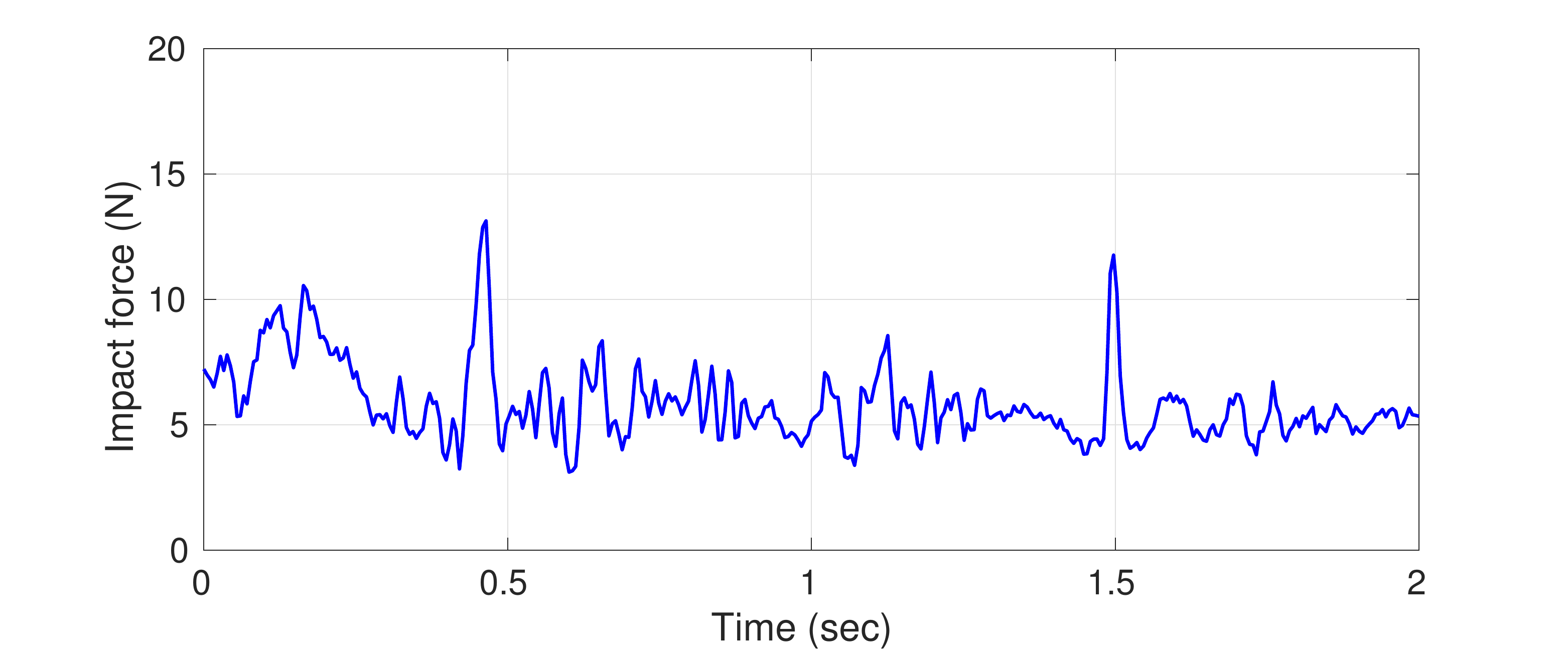}
\caption{\label{Fig:RealGrasp3} }
\end{subfigure}
\caption{Impact force evolutions for three different grasps, performed with a real robot colliding with an obstacle (water-filled bottle). The point of collision is clearly visible on the plots, in the form of a sharp peak in the force measurements. Once again, different grasps lead to different impact forces (observed in the first large peak in contact force, just before 0.5 seconds), consistent with the simulation experiments. The magnitude of the peak collision forces varies from 13N (safest grasp) to 17N (most dangerous grasp), i.e. roughly 30\% difference in safety for even this very simple case, with seemingly modest changes in the choice of grasp pose. Note that the contact force fluctuations, after the collision, are most likely attributable to gravity torque compensation behaviour of the real Baxter robot, combined with minor inertial effects (note that the bottle obstacle has toppled over at this point, so no obstacle exists to account for this post-impact fluctuation behaviour). (\subref{Fig:RealGrasp1}) First grasp. (\subref{Fig:RealGrasp2}) Second grasp. (\subref{Fig:RealGrasp3}) Third grasp.}  
\label{fig:RealExpResults}
\end{figure}

In the simulation experiments, we were able to create a virtual force sensor to measure impact forces. In future work, we hope to acquire an expensive, high-precision, force-torque sensor for accurately measuring these forces in experiments with real robots. Since such a sensor was unavailable to us at the time of writing, we used the Baxter robot's (rather noisy) force estimation system, based on a model of its series-elastic actuators combined with joint rotation sensors, to approximately measure the forces experienced by its end-effector. The results are shown in Fig.~\ref{fig:RealExpResults}.

% This allows us to avoid any fault due to the hard collision. However, based on the theory of the impact force we can use the force as representative as collision with a hard constrain. 
%This confirms that our choice of our set-up in the real experiment does not violate our assumptions and we can trust to the force signal at collision time measured from the sensor of the Baxter. 
% We keep the grasping points identical to the simulation of book experiment. 
% The robot approached the object, grasped it from the specified grasping point and executed the task. 

The time of collision can be clearly seen in Fig.~\ref{fig:RealExpResults} as a large peak in the contact force, shortly before 0.5 seconds after motion begins. From Fig.~\ref{fig:RealExpResults} it can be seen that the maximum force during impact is different for each grasp, with around around 30\% difference between safest grasp and least safe grasp. This is consistent with the simulation experiment, where the minimum impact force (safest grasp) is around 100N, while the maximum impact force (least safe grasp) is around 130N.

Note that, in the simulation experiment, the overall magnitude of contact forces was higher, because the robot was colliding with a completely rigid and static obstacle. In contrast, in the real robot experiments, the robot collides with a light-weight, deformable obstacle (full water bottle), which topples over after impact. Nevertheless, the differences in impact force between the safest and least safe grasps were similar in both real and simulation experiments, with approximately 30\% difference in both cases. In both real and simulation experiments, the grasp with minimum effective mass results in minimum impact force. In both experiments, the most safe grasp was located on the right side of the book (Fig.~\ref{Fig:G3}) and the least safe grasp was on the left side (Fig.~\ref{Fig:G1}).

\section{Discussion}
The results acquired from all experiments support our proposed methodology. As demonstrated, selection of an optimal grasp can reduce the impact in case of a collision. Grasps that minimise the computed effective mass along a desired post-grasp trajectory, are an effective predictor of grasps that maximise safety with respect to post-grasp collisions. This paper has proposed a way of pre-calculating this optimality, by using the effective mass of the augmented dynamics, integrated over the desired post-grasp trajectory. As seen in Fig.~\ref{Fig:allGraspsImage} and Fig.~\ref{fig:BookGrasps}, the difference between effective mass values can vary significantly between different grasp choices. Furthermore, effective mass differences, between different grasps, result in significantly different impact forces in post-grasp collisions.

Our experiments were designed to be simple and able to isolate and measure the effect of each grasp at the instance of collision. As we were interested primarily in minimizing the impact force, we chose not to investigate, or optimise for, post-impact phenomena such as manipulator stability after impact, or steady-state response of the contact force. The connection of these phenomena to post-grasp manipulation and the grasped object's inertial properties can serve as an interesting topic for future research.  Other ways of potentially enhancing our method in future include further minimization of the force by redundancy control, or evaluating how changes in the object's inertia tensor during the task, due to e.g. grasp slippage, would perform.

It should also be noted that the dynamics of the robot may not always be available in analytical form, and may need to be identified. The same can be said for the inertial properties of grasped objects, where only approximations can be given, with some margin of error, in many real-world applications. Nevertheless, our proposed methodology is suitable for providing safe grasps, even when using such approximations.

\section{Conclusion}
In this paper, we provided empirical evidence to support our assertion that different choices of grasp lead to different degrees of safety, w.r.t. collisions during post-grasp manipulations of grasped objects. We showed how different grasp choices can result in different augmented dynamics, and different effective masses at successive robot configurations along a desired post-grasp trajectory.

We designed experiments specifically to fix all experimental variables other than the choice of grasp location, so as to isolate and demonstrate how changes in the impact force can be attributed only to the change in grasp location. We implemented: physics engine simulations with a highly controllable virtual tensor object; further simulation experiments with a book object; as well as conducting experiments with a real robot and real book object. In all cases, the experimental results consistently support our thesis that: i) selection of grasp location can make a robot significantly safer; ii) pre-calculations of the effective mass, along a desired post-grasp trajectory, are a useful way of identifying the grasp choices that minimise the severity of post-grasp collisions.

This paper represents only initial work in this field. Future work could usefully include:
\begin{itemize}
\item repeating the experiments with heavy-duty industrial manipulators, where higher payloads can be supported so that more accurate impact measurements can be acquired;
\item combining the post-grasp collision safety metric, proposed in this paper, with ``graspability'' or ``grasp likelihood'' metrics, obtained from state-of-the-art grasp planners, in order to mutually optimise grasp safety with grasp stability;
\item combining the ideas proposed in this paper with other methods for estimating the mass and mass distribution of an object;
\item combining the ideas of this paper with our previous works \cite{Mavrakis2016,Ghalamzan2016} on task-relevant grasp planning, to provide a series of criteria for achieving grasp selection via multi-objective optimisation.
\end{itemize}

\bibliographystyle{ieeetr}
\bibliography{ref}

\end{document}